\pdfoutput=1
\documentclass[accepted]{uai2022} 

\usepackage[american]{babel}

\usepackage[round]{natbib} 
\usepackage{mathtools} 
\usepackage{booktabs} 
\usepackage{zref-xr}
\usepackage{nameref}
\usepackage{hyperref}
\usepackage{caption}
\usepackage{subcaption}
\usepackage{bbm}
\usepackage{xifthen}
\usepackage{graphicx}
\usepackage{adjustbox}

\usepackage{standalone}
\usepackage[dvipsnames]{xcolor}
\usepackage{tikz}
\usetikzlibrary{positioning}
\usetikzlibrary{arrows}
\usetikzlibrary{calc,fit}
\usetikzlibrary{shapes.geometric}
\usetikzlibrary{shapes.misc}
\usetikzlibrary{decorations.pathmorphing}
\usetikzlibrary{decorations.pathreplacing}
\usetikzlibrary{snakes}
\usepackage{pgfplots}
\usepgfplotslibrary{groupplots}
\pgfplotsset{compat=1.16}
\usepgfplotslibrary{statistics}

\usepackage{amsmath}
\usepackage{amssymb}
\usepackage{mathtools}
\usepackage{amsthm}

\definecolor{darkblue}{rgb}{0.0, 0.0, 0.55}
\hypersetup{
	pdftitle={Bayesian Structure Learning with Generative Flow Networks},
	pdfkeywords={},
	pdfborder=0 0 0,
	pdfpagemode=UseNone,
	colorlinks=true,
	linkcolor=darkblue, 
	citecolor=darkblue, 
	filecolor=darkblue, 
	urlcolor=darkblue, 
	pdfview=FitH,
	pdfauthor={Tristan Deleu, Antonio Gois, Chris Emezue, Mansi Rankawat, Simon Lacoste-Julien, Stefan Bauer, Yoshua Bengio}
}

\usepackage[capitalize,noabbrev]{cleveref}

\theoremstyle{plain}

\theoremstyle{definition}

\theoremstyle{remark}

\usepackage{amsmath,amsfonts,bm}

\def\eqref#1{equation~\ref{#1}}

\def\1{\bm{1}}

\def\vm{{\bm{m}}}

\def\vx{{\bm{x}}}

\DeclareMathAlphabet{\mathsfit}{\encodingdefault}{\sfdefault}{m}{sl}
\SetMathAlphabet{\mathsfit}{bold}{\encodingdefault}{\sfdefault}{bx}{n}

\def\gD{{\mathcal{D}}}

\def\gL{{\mathcal{L}}}

\def\gN{{\mathcal{N}}}

\def\gS{{\mathcal{S}}}

\def\gX{{\mathcal{X}}}

\newcommand{\E}{\mathbb{E}}

\crefformat{section}{#2Section~#1#3}
\crefformat{appendix}{#2Appendix~#1#3}

\crefformat{equation}{(#2#1#3)}
\crefmultiformat{equation}{#2Equations~#1#3}%
{ \&~#2#1#3}{, #2#1#3}{, \&~#2#1#3}

\crefformat{table}{#2Table~#1#3}
\crefformat{figure}{#2Figure~#1#3}

\crefformat{theorem}{#2Theorem~#1#3}
\crefmultiformat{theorem}{#2Theorems~#1#3}%
{ \&~#2#1#3}{, #2#1#3}{, and~(#2#1#3)}

\crefformat{lemma}{#2Lemma~#1#3}
\crefformat{proposition}{#2Proposition~#1#3}
\crefmultiformat{proposition}{#2Propositions~#1#3}%
{ \&~#2#1#3}{, #2#1#3}{, and~(#2#1#3)}

\crefformat{algorithm}{#2Algorithm~#1#3}
\crefmultiformat{algorithm}{#2Algorithms~#1#3}%
{ \&~#2#1#3}{, #2#1#3}{, and~(#2#1#3)}

\crefformat{corollary}{#2Corollary~#1#3}
\crefformat{definition}{#2Definition~#1#3}

\crefformat{assumption}{#2Assumption~#1#3}
\crefmultiformat{assumption}{#2Assumptions~#1#3}%
{ \&~#2#1#3}{, #2#1#3}{, and~(#2#1#3)}

\newcommand{\children}{\mathrm{Ch}}
\newcommand{\Pa}{\mathrm{Pa}}

\title{Bayesian Structure Learning with Generative Flow Networks}

\makeatletter
\def\thanks#1{\protected@xdef\@thanks{\@thanks\protect\footnotetext{#1}}}
\makeatother

\makeatletter
\renewcommand\AB@affilsepx{\hspace{1em}\protect\Affilfont}
\makeatother

\author[1]{Tristan~Deleu\thanks{Correspondence to: Tristan Deleu <\href{mailto:deleutri@mila.quebec}{deleutri@mila.quebec}>}}
\author[1]{Ant\'{o}nio~G\'{o}is}
\author[2,*]{Chris~Emezue\thanks{\textsuperscript{*}Work done during an internship at Mila}}
\author[1]{Mansi~Rankawat}
\author[1,4]{\authorcr{Simon~Lacoste-Julien}}
\author[3,5]{Stefan~Bauer}
\author[1,4,6]{Yoshua~Bengio}
\affil[1]{Mila, Universit\'{e} de Montr\'{e}al}
\affil[2]{Technical University of Munich}
\affil[3]{KTH Stockholm\protect\\[0.5em]}
\affil[4]{CIFAR AI Chair}
\affil[5]{CIFAR Azrieli Global Scholar}
\affil[6]{CIFAR Senior Fellow}
  
\begin{document}
\maketitle

\begin{abstract}
In Bayesian structure learning, we are interested in inferring a distribution over the directed acyclic graph (DAG) structure of Bayesian networks, from data. Defining such a distribution is very challenging, due to the combinatorially large sample space, and approximations based on MCMC are often required. Recently, a novel class of probabilistic models, called Generative Flow Networks (GFlowNets), have been introduced as a general framework for generative modeling of discrete and composite objects, such as graphs. In this work, we propose to use a GFlowNet as an alternative to MCMC for approximating the posterior distribution over the structure of Bayesian networks, given a dataset of observations. Generating a sample DAG from this approximate distribution is viewed as a sequential decision problem, where the graph is constructed one edge at a time, based on learned transition probabilities. Through evaluation on both simulated and real data, we show that our approach, called DAG-GFlowNet, provides an accurate approximation of the posterior over DAGs, and it compares favorably against other methods based on MCMC or variational inference.
\end{abstract}

\section{Introduction}
\label{sec:introduction}
Bayesian networks \citep{pearl1988bayesiannetworks} are a popular framework of choice for representing uncertainty about the world in multiple scientific domains, including medical diagnosis \citep{lauritzen1988munin,heckerman1992pathfinder}, molecular biology \citep{friedman2004cellular,sebastiani2005genetic}, and ecological modeling \citep{varis1999environmental,marcot2006guidelinesecological}. For many applications, the structure of the Bayesian network, represented as a directed acyclic graph (DAG) and encoding the statistical dependencies between the variables of interest, is assumed to be known based on knowledge from domain experts. However, when this graph is unknown, we can learn the DAG structure of the Bayesian network from data alone in order to discover these statistical (or possibly causal) relationships. This may form the basis of novel scientific theories.

Given a dataset of observations, most of the existing algorithms for structure learning return a single DAG (or a single equivalence class; \citealp{chickering2002ges}), and in practice those may lead to poorly calibrated predictions \citep{madigan1994enhancing}, especially in cases where data is limited. Instead of learning a single graph candidate, we can view the problem of structure learning from a Bayesian perspective and infer the posterior over graphs $P(G\mid \gD)$, given a dataset of observations $\gD$, to account for the epistemic uncertainty over models. Except in limited settings \citep{koivisto2006exact,meilua2006tree}, characterizing a whole distribution over DAGs remains intractable because of its combinatorially large sample space and the complex acyclicity constraint. Therefore, we must often resort to approximations of this posterior distribution, e.g., based on MCMC or, more recently, variational~inference.

In this paper, we propose to use a novel class of probabilistic models called \emph{Generative Flow Networks} \citep[GFlowNets;][]{bengio2021gflownet,bengio2021gflownetfoundations} to approximate this posterior distribution over DAGs. A GFlowNet is a generative model over discrete and composite objects that treats the generation of a sample as a sequential decision problem. This makes it particularly appealing for modeling a distribution over graphs, where sample graphs are constructed sequentially, starting from the empty graph, by adding one edge at a time. In the context of Bayesian structure learning, we also introduce improvements over the original GFlowNet framework, including a novel flow-matching condition and corresponding loss function, a hierarchical probabilistic model for forward transitions, and using additional tools from the literature on Reinforcement Learning (RL). We call our method \emph{DAG-GFlowNet}, to emphasize that the support of the distribution induced by the GFlowNet is exactly the space of DAGs, unlike some variational approaches that may sample cyclic graphs \citep{annadani2021vcn,lorch2021dibs}. Compared to MCMC, which operates through local moves in the sample space (here, adding or removing edges of a graph) and is therefore subject to slow mixing \citep{friedman2003ordermcmc}, DAG-GFlowNet yields a sampling process that samples iid. DAGs, \mbox{each of them constructed from
 scratch}.

We evaluate DAG-GFlowNet on various problems with simulated and real data, on both discrete and linear-Gaussian Bayesian networks. Furthermore, we show that DAG-GFlowNet can be applied on both observational and interventional data, by modifying standard Bayesian scores \citep{cooper1999interventional}. On smaller graphs, we also show that it is capable of learning an accurate approximation of the exact posterior distribution. The code is available \mbox{online}.\footnote{\href{https://github.com/tristandeleu/jax-dag-gflownet}{\scriptsize \texttt{https://github.com/tristandeleu/jax-dag-gflownet}}}

\section{Related Work}
\label{sec:related-work}
\paragraph{Markov chain Monte Carlo} Methods based on MCMC have been particularly popular in Bayesian structure learning to approximate the posterior distribution. Structure MCMC \citep[MC\textsuperscript{3};][]{madigan1995structuremcmc} simulates a Markov chain in the space of DAGs, through local moves (e.g. adding or removing an edge). Working directly with DAGs leads to slow mixing though; to improve mixing, \citet{friedman2003ordermcmc} proposed a sampler in the space of node orders, that introduced a bias \citep{ellis2008bias}. This was further refined by either modifying the underlying space of the Markov chain \citep{kuipers2017partitionmcmc,niinimaki2016partialordermcmc}, or its local moves \citep{mansinghka2006gibbs,eaton2007bayesian,kuipers2021efficient}. Recently, \citet{viinikka2020gadget} incorporated many of these advances into an efficient MCMC sampler called \textit{Gadget}.

\paragraph{Variational Inference} In the context of structure learning, applying the recent advances in approximate inference based on gradient methods can be difficult due to the discrete nature of the problem \citep{lorch2021dibs}. \citet{cundy2021bcdnets} decomposed the adjacency matrix of a DAG into a triangular matrix and a permutation, and used a continuous relaxation to parametrize a distribution over permutations. Other methods \citep{annadani2021vcn,lorch2021dibs} encode the acyclicity constraint into a soft prior $P(G)$, based on continuous characterizations of acyclicity \citep{zheng2018notears}. While the effect of this prior can be made arbitrarily strong, this does not guarantee that the graphs sampled from the resulting distribution are acyclic. By contrast, our approach guarantees by construction that the support of the posterior approximation is exactly the space of DAGs.

\paragraph{Sequential decisions} In this work, we treat the construction of a sample graph from the posterior as a sequential decision problem, starting from the empty graph and adding one edge at a time. \citet{li2018gengraphs} use a similar process for creating a generative model over graphs with a fixed ordering over nodes. Similarly, although they do not consider a distribution over graphs, \citet{buesing2020approximate} use a variant of Monte Carlo Tree Search to approximate a distribution over a pre-specified ordering of discrete random variables. Our method, based on Generative Flow Networks, does not make any assumption on the order in which the edges are added, and multiple edge insertion sequences may lead to the same DAG. \citet{zhu2020causaldiscoveryrl} learn a single high-scoring structure using RL; however, unlike our approach, the creation of this graph does not involve sequential decisions.

\section{Background}
\label{sec:background}
A \emph{Bayesian network} is a probabilistic model over $d$ random variables $\{X_{1}, \ldots, X_{d}\}$, whose joint distribution factorizes according to a DAG $G$ as
\begin{equation}
    P(X_{1}, \ldots, X_{d}) = \prod_{k=1}^{d}P\big(X_{k} \mid \Pa_{G}(X_{k})\big),
\end{equation}
where $\Pa_{G}(X)$ is the set of parents of node $X$ in $G$. Similarly, we denote by $\children_{G}(X)$ the children of $X$; when the context is clear, we may drop the explicit dependency on $G$.

\subsection{Generative Flow Networks}
\label{sec:generative-flow-networks}
Originally introduced to encourage the discovery of diverse modes of an unnormalized distribution \citep{bengio2021gflownet}, \emph{Generative Flow Networks} \citep[GFlowNets;][]{bengio2021gflownetfoundations} are a class of generative models over a discrete and structured sample space $\gX$. The structure of a GFlowNet is defined by a DAG over some states $s \in \gS$; in general, the sample space over which we wish to define a distribution is only a subset of the overall state space of the GFlowNet: $\gX \subseteq \gS$. Samples $s \in \gX$ are constructed sequentially by following the edges of the DAG, starting from a fixed initial state $s_{0}$. We also define a special absorbing state $s_{f}$, called the terminal state, indicating when the sequential construction terminates; some of the states $s \in \gX$ are connected to $s_{f}$, and we call them \emph{complete states}.\footnote{``Complete'' here means that the state is a valid sample from the distribution induced by the GFlowNet. This must not be confused with a ``complete graph'', where all the nodes are connected to one another, when the states are DAGs (see \cref{sec:gflownet-over-dags}).} For example, \citet{bengio2021gflownet} use a GFlowNet to define a distribution over molecules, where $\gX$ would correspond to the space of all (complete) molecules, which are constructed piece by piece by attaching a new block to an atom in a possibly partially constructed molecule (i.e. a state in $\gS \backslash \gX$). Another example of a GFlowNet structure is given in Fig.~\ref{fig:gflownet-dags}, illustrating the sequential process of constructing a DAG. A GFlowNet is structurally equivalent to a Markov Decision Process \citep[MDP;][]{puterman1994mdp} with deterministic dynamics, or a Markov Reward Process \citep{howard1971mrp}.

In addition to the DAG structure over states, every complete state $s \in \gX$ is associated with a \emph{reward} $R(s) \geq 0$, indicating a notion of ``preference'' for certain states. By convention, $R(s) = 0$ for any incomplete state $s \in \gS \backslash \gX$. The goal of a GFlowNet is to find a \emph{flow} that satisfies, for all states $s' \in \gS$, the following \emph{flow-matching condition}:
\begin{equation}
    \sum_{\mathclap{s \in \Pa(s')}}\;F_{\theta}(s \rightarrow s') -\;\sum_{\mathclap{s'' \in \children(s')}}\;F_{\theta}(s' \rightarrow s'') = R(s'),
    \label{eq:flow-matching-condition}
\end{equation}
where $F_{\theta}(s \rightarrow s') \geq 0$ is a scalar representing the flow from state $s$ to $s'$, typically parametrized by a neural network. Putting it in words, the overall flow going into $s'$ is equal to the flow going out of $s'$, plus some residual $R(s')$. To learn the parameters $\theta$ of the flow with SGD, we can turn \cref{eq:flow-matching-condition} into a regression problem, e.g. using a least squares objective over sampled states.

If the conditions in \cref{eq:flow-matching-condition} are satisfied for all states $s'$, a GFlowNet induces a generative process to sample complete states $s \in \gX$ with probability proportional to $R(s)$. Starting from the initial state $s_{0}$, if we sample a complete trajectory $(s_{0}, s_{1}, \ldots, s_{T}, s, s_{f})$ using the transition probability defined as the normalized outgoing flow
\begin{equation}
    P(s_{t+1} \mid s_{t}) \propto F_{\theta}(s_{t} \rightarrow s_{t+1}),
    \label{eq:forward-transition-probability}
\end{equation}
with the conventions $s_{T+1} = s$ and $s_{T+2} = s_{f}$, then $s$ is sampled with probability $P(s) \propto R(s)$. Note that the linear system in \cref{eq:flow-matching-condition} is in general underdetermined, and therefore it may admit many solutions $F_{\theta}(s \rightarrow s')$ that all induce the same distribution $\propto R(s)$. Unlike MCMC, each sample $s \in \gX$ is constructed from scratch, starting at the initial state $s_{0}$, instead of traversing $\gX$ from sample to sample. Therefore, the underlying Markov process of the GFlowNet does not have to be irreducible, which is typically necessary in MCMC, but merely requires all the complete states to be reachable from the initial state. Finally, although GFlowNets borrow terminology from RL and control theory (e.g. MDPs, rewards, trajectories), their objective is different from the typical RL training objective: the latter seeks to maximize a function of the rewards, while the goal of GFlowNets is to model the whole distribution proportional to the rewards.

\subsection{Detailed-balance condition}
\label{sec:detailed-balance-condition}
Since the flows are added together, one of the downsides of the flow-matching condition is that flows tend to be orders of magnitude larger the closer we are of the initial state \citep{bengio2021gflownet}, making it challenging to parametrize $F_{\theta}$. \citet{bengio2021gflownetfoundations} proposed an alternative characterization of GFlowNets inspired by the detailed-balance equations from the literature on Markov chains \citep{grimmett2020probability}. Instead of working with flows, this condition uses a parametrization of the forward transition probability $P_{\theta}(s_{t+1} \mid s_{t})$ directly, together with a backward transition probability $P_{B}(s_{t} \mid s_{t+1})$ to enforce reversibility. As opposed to $P_{\theta}(s_{t+1} \mid s_{t})$, which is a distribution over the children of $s_{t}$, $P_{B}(s_{t} \mid s_{t+1})$ is a distribution over the parents of $s_{t+1}$ in the structure of the GFlowNet. If all the states of the GFlowNet are complete (except the terminal state $s_{f}$), which will be the case here for generating DAGs, then we show in \cref{app:detailed-balance-condition} that we can write the \emph{detailed-balance condition} for all transitions $s \rightarrow s'$ as follows:
\begin{equation*}
    R(s')P_{B}(s\mid s')P_{\theta}(s_{f}\mid s) = R(s)P_{\theta}(s'\mid s)P_{\theta}(s_{f}\mid s').
    \label{eq:detailed-balance-condition}
\end{equation*}
Similar to \cref{sec:generative-flow-networks}, finding $P_{\theta}$ and $P_{B}$ that satisfy this condition for all the transitions $s \rightarrow s'$ of the GFlowNet also yields a sampling process of complete states $s$ with probability proportional to $R(s)$, based on the forward transition probability $P_{\theta}(s_{t+1}\mid s_{t})$. Because this system of equations also admits many solutions, similar to \cref{eq:flow-matching-condition}, we can set the backward transition probability $P_{B}$ to some fixed distribution (e.g. the uniform distribution over the parent states) to reduce the search space, making $P_{\theta}$ the only quantity to learn and, with enough capacity (to satisfy the constraints), there is a unique solution $P_{\theta}$ \citep{bengio2021gflownetfoundations}.

To fit the parameters $\theta$ of the forward transition probability, we can minimize the following non-linear least squares objective for all the transitions $s \rightarrow s'$ of the GFlowNet, called the \emph{detailed-balance loss}:
\begin{equation}
    \gL(\theta) = \sum_{s \rightarrow s'}\bigg[\log \frac{R(s')P_{B}(s\mid s')P_{\theta}(s_{f}\mid s)}{R(s)P_{\theta}(s'\mid s)P_{\theta}(s_{f}\mid s')}\bigg]^{2}.
    \label{eq:detailed-balance-loss}
\end{equation}
Alternatively, we can minimize this loss in expectation, using a distribution $\pi(s \rightarrow s')$ with full support over transitions (i.e. for all transitions $s \rightarrow s'$ in the GFlowNet, we have $\pi(s \rightarrow s') > 0$; see \cref{sec:off-policy-learning}).

\section{GFlowNet over Directed Acyclic Graphs}
\label{sec:gflownet-over-dags}
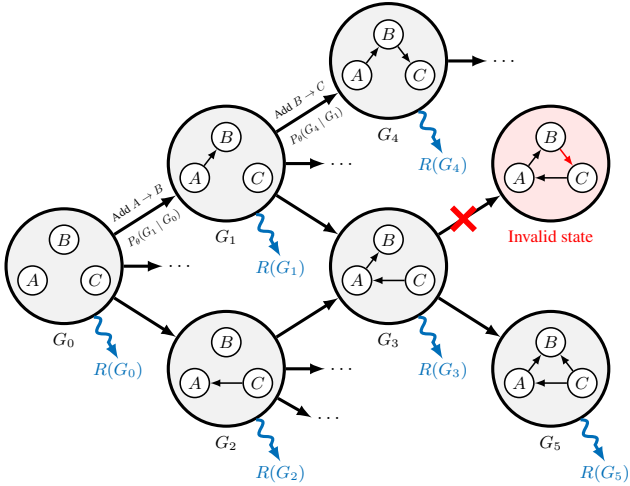
\begin{figure}[t!]
    \centering
    \begin{tikzpicture}[every node/.style={inner sep=0pt}, var/.style={draw=black, minimum size=15pt, circle, font=\small, thick, anchor=center, fill=white}, factor/.style={fill=black, minimum size=5pt, rectangle, anchor=center}, edge/.style={thick, -latex}, modulation/.style={thick, -|}, zlabel/.style={font=\scriptsize, yshift=-2pt}, observed/.style={fill=gray!30}, flowlabel/.style={midway, font=\scriptsize}, rewardlabel/.style={pos=1, font=\scriptsize, right}, gfn_edge/.style={ultra thick, -latex}, gfn_node/.style={ultra thick, circle, draw, text height=3.5em, fill=gray!10}, gfn_other/.style={pos=1, anchor=west, inner sep=2pt}, adjacency/.style={minimum width=10pt, minimum height=10pt, rectangle, rounded corners=2pt, font=\bfseries\scriptsize}, y=12pt, x=50pt]

\node[] (gflownet) at (0, 0) {\begin{tikzpicture}

\node[gfn_node, label={[font=\small, inner sep=4pt]270:$G_{0}$}] (graph1) at (0, 0) {\begin{tikzpicture}
    \node[var] (A) {$A$};
    \node[var, above right=1em and 0.5em of A] (B) {$B$};
    \node[var, below right=1em and 0.5em of B] (C) {$C$};
    \node[below=0.1em of A] (D) {};
\end{tikzpicture}};

\node[gfn_node, above right=1em and 4em of graph1, label={[font=\small, inner sep=4pt]270:$G_{1}$}] (graph2) {\begin{tikzpicture}
    \node[var] (A) {$A$};
    \node[var, above right=1em and 0.5em of A] (B) {$B$};
    \node[var, below right=1em and 0.5em of B] (C) {$C$};
    \node[below=0.1em of A] (D) {};
    
    \draw[edge] (A) -- (B);
\end{tikzpicture}};

\node[gfn_node, below right=1em and 4em of graph1, label={[font=\small, inner sep=4pt]270:$G_{2}$}] (graph3) {\begin{tikzpicture}
    \node[var] (A) {$A$};
    \node[var, above right=1em and 0.5em of A] (B) {$B$};
    \node[var, below right=1em and 0.5em of B] (C) {$C$};
    \node[below=0.1em of A] (D) {};
    
    \draw[edge] (C) -- (A);
\end{tikzpicture}};

\node[gfn_node, below right=1em and 4em of graph2, label={[font=\small, inner sep=4pt]270:$G_{3}$}] (graph4) {\begin{tikzpicture}
    \node[var] (A) {$A$};
    \node[var, above right=1em and 0.5em of A] (B) {$B$};
    \node[var, below right=1em and 0.5em of B] (C) {$C$};
    \node[below=0.1em of A] (D) {};
    
    \draw[edge] (A) -- (B);
    \draw[edge] (C) -- (A);
\end{tikzpicture}};

\node[gfn_node, above right=1em and 4em of graph2, label={[font=\small, inner sep=4pt]270:$G_{4}$}] (graph5) {\begin{tikzpicture}
    \node[var] (A) {$A$};
    \node[var, above right=1em and 0.5em of A] (B) {$B$};
    \node[var, below right=1em and 0.5em of B] (C) {$C$};
    \node[below=0.1em of A] (D) {};
    
    \draw[edge] (A) -- (B);
    \draw[edge] (B) -- (C);
\end{tikzpicture}};

\node[gfn_node, above right=1em and 4em of graph4, fill=red!10, label={[font=\small, inner sep=4pt, red]270:Invalid state}] (graph6) {\begin{tikzpicture}
    \node[var] (A) {$A$};
    \node[var, above right=1em and 0.5em of A] (B) {$B$};
    \node[var, below right=1em and 0.5em of B] (C) {$C$};
    \node[below=0.1em of A] (D) {};
    
    \draw[edge] (A) -- (B);
    \draw[edge] (C) -- (A);
    \draw[edge, red] (B) -- (C);
\end{tikzpicture}};

\node[gfn_node, below right=1em and 4em of graph4, label={[font=\small, inner sep=4pt]270:$G_{5}$}] (graph7) {\begin{tikzpicture}
    \node[var] (A) {$A$};
    \node[var, above right=1em and 0.5em of A] (B) {$B$};
    \node[var, below right=1em and 0.5em of B] (C) {$C$};
    \node[below=0.1em of A] (D) {};
    
    \draw[edge] (A) -- (B);
    \draw[edge] (C) -- (A);
    \draw[edge] (C) -- (B);
\end{tikzpicture}};

\draw[gfn_edge] (graph1) -- (graph2) node[midway, sloped, above=5pt, scale=0.6] {Add $A \rightarrow B$} node[midway, sloped, below=5pt, scale=0.6] {$P_{\theta}(G_{1} \mid G_{0})$};
\draw[gfn_edge] (graph1) -- (graph3);
\draw[gfn_edge] (graph2) -- (graph4);
\draw[gfn_edge] (graph3) -- (graph4);
\draw[gfn_edge] (graph2) -- (graph5) node[midway, sloped, above=5pt, scale=0.6] {Add $B \rightarrow C$} node[midway, sloped, below=5pt, scale=0.6] {$P_{\theta}(G_{4} \mid G_{1})$};
\draw[gfn_edge] (graph4) -- (graph6) node[circle, minimum size=1.5em, pos=0.4, draw=none] (cross) {};
\draw[gfn_edge] (graph4) -- (graph7);

\draw[gfn_edge] (graph1.0) -- ++(0:2em) node[gfn_other] {$\ldots$};
\draw[gfn_edge] (graph2.0) -- ++(0:2em) node[gfn_other] {$\ldots$};
\draw[gfn_edge] (graph3.0) -- ++(0:2em) node[gfn_other] {$\ldots$};
\draw[gfn_edge] (graph3.330) -- ++(330:2em) node[gfn_other] {$\ldots$};
\draw[gfn_edge] (graph5.0) -- ++(0:2em) node[gfn_other] {$\ldots$};

\draw[line width=3pt, red] (cross.45) -- (cross.225);
\draw[line width=3pt, red] (cross.135) -- (cross.315);

\draw[ultra thick, NavyBlue, snake=coil, segment aspect=0, line after snake=1pt] (graph1.300) -- ++(300:2em) [snake=none, ultra thick, -latex, NavyBlue] -- ++(300:5pt) node[pos=1, anchor=north, inner sep=1pt, NavyBlue, font=\small] {$R(G_{0})$};
\draw[ultra thick, NavyBlue, snake=coil, segment aspect=0, line after snake=1pt] (graph2.300) -- ++(300:2em) [snake=none, ultra thick, -latex, NavyBlue] -- ++(300:5pt) node[pos=1, anchor=north, inner sep=1pt, NavyBlue, font=\small] {$R(G_{1})$};
\draw[ultra thick, NavyBlue, snake=coil, segment aspect=0, line after snake=1pt] (graph3.300) -- ++(300:2em) [snake=none, ultra thick, -latex, NavyBlue] -- ++(300:5pt) node[pos=1, anchor=north, inner sep=1pt, NavyBlue, font=\small] {$R(G_{2})$};
\draw[ultra thick, NavyBlue, snake=coil, segment aspect=0, line after snake=1pt] (graph4.300) -- ++(300:2em) [snake=none, ultra thick, -latex, NavyBlue] -- ++(300:5pt) node[pos=1, anchor=north, inner sep=1pt, NavyBlue, font=\small] {$R(G_{3})$};
\draw[ultra thick, NavyBlue, snake=coil, segment aspect=0, line after snake=1pt] (graph5.300) -- ++(300:2em) [snake=none, ultra thick, -latex, NavyBlue] -- ++(300:5pt) node[pos=1, anchor=north, inner sep=1pt, NavyBlue, font=\small] {$R(G_{4})$};
\draw[ultra thick, NavyBlue, snake=coil, segment aspect=0, line after snake=1pt] (graph7.300) -- ++(300:2em) [snake=none, ultra thick, -latex, NavyBlue] -- ++(300:5pt) node[pos=1, anchor=north, inner sep=1pt, NavyBlue, font=\small] {$R(G_{5})$};

\end{tikzpicture}};

\end{tikzpicture}
    \vspace*{-1.5em}
    \caption{Structure of a GFlowNet over DAGs. The states of the GFlowNet correspond to DAGs, with the initial state $G_{0}$ being the completely disconnected graph. Each state $G$ is complete (i.e. connected to the terminal state $s_{f}$, represented by blue arrows for brevity) and associated to a reward $R(G)$. Transitioning from one state to another corresponds to adding an edge to the graph. The state in red is invalid since the graph includes a cycle.}
    \label{fig:gflownet-dags}
    \vspace*{-1em}
\end{figure}
Our objective in this paper is to construct a distribution over DAGs. This is a challenging problem in general, as the space of DAGs is discrete and combinatorially large. We propose to use a GFlowNet to model such a distribution; this is particularly appropriate here since graphs are composite objects, and the acyclicity constraint can be obtained by constraining the allowed actions in each state (as in \cref{fig:gflownet-dags}). Note that the DAGs in this section and thereafter represent the states of the GFlowNet, and they must not be confused with the DAG structure of the GFlowNet itself.

\subsection{Structure of the GFlowNet}
\label{sec:structure-gflownet}
We consider a GFlowNet where the states are DAGs over $d$ (labeled) nodes. Since the states of the GFlowNet are graphs, we will use the notation $G$ to denote a state, in favour of $s$ as in \cref{sec:generative-flow-networks}, except for the terminal state $s_{f}$. A transition $G \rightarrow G'$ in this GFlowNet corresponds to adding an edge to $G$ to obtain the graph $G'$; in other words, the graphs are constructed one edge at a time, starting from the initial state $G_{0}$, which is the fully disconnected graph over $d$ nodes. Since we assume that all the states $G$ of the GFlowNet are valid DAGs, they are all complete (i.e. connected to the terminal state $s_{f}$) with a corresponding reward $R(G)$. \cref{fig:gflownet-dags} shows an illustration of the structure of such a GFlowNet, where the states are DAGs over $d=3$ nodes. This application to graphs highlights the importance of the DAG structure of the GFlowNet, since there can be multiple paths leading to the same state: for any graph $G$ with $k$ edges, there are $k!$ possible paths from $G_{0}$ leading to $G$, because the edges of $G$ may have been added in any order.

To guarantee the integrity of the GFlowNet, we have to ensure that adding a new edge to some state $G$ also yields a valid DAG, meaning that this edge (1) must not be already present in $G$, and (2) must not introduce a cycle. Fortunately, we can filter out invalid actions using some mask $\vm$ associated to the graph, built from the adjacency matrix of $G$ and the transitive closure of its transpose, and that can be updated efficiently after the addition an edge \citep{giudici2003improvingmcmc}. A description of this update is given in \cref{app:mask} for completeness.

\subsection{Forward transition probabilities}
\label{sec:forward-transition-probabilities}
Following \cref{sec:detailed-balance-condition}, the GFlowNet may be parametrized only by the forward transition probabilities $P_{\theta}(G_{t+1} \mid G_{t})$; here, $G_{t+1}$ might be the terminal state $s_{f}$ by abuse of notation. To make sure that the detailed-balance conditions can be satisfied, we need to define these transition probabilities using a sufficiently expressive function, such as a neural network. We use a hierarchical model, where the forward transition probabilities are defined using two neural networks: (1) a network modeling the probability of terminating $P_{\theta}(s_{f} \mid G)$, and (2) another giving the probability $P_{\theta}(G' \mid G, \neg s_{f})$ of transitioning to a new graph $G'$, given that we do not terminate. The probability of taking a transition $G \rightarrow G'$ is then given by
\begin{equation}
    P_{\theta}(G' \mid G) = \big(1 - P_{\theta}(s_{f} \mid G)\big) P_{\theta}(G' \mid G, \neg s_{f}).
    \label{eq:transition-proba-hierarchical}
\end{equation}
In practice, as $G'$ is the result of adding an edge to the DAG $G$, we can model $P_{\theta}(G' \mid G, \neg s_{f})$ as a probability distribution over the $d^{2}$ possible edges one could add to $G$---this includes self-loops, for simplicity, even though these actions are guaranteed to be invalid. We can use the mask $\vm$ introduced in \cref{sec:structure-gflownet} to filter out actions that would not lead to a valid DAG $G'$ and set $P_{\theta}(G' \mid G, \neg s_{f}) = 0$ for any invalid action (as well as normalize $P_{\theta}$ accordingly).

\subsection{Parametrization with Linear Transformers}
\label{sec:parametrization-linear-transformers}
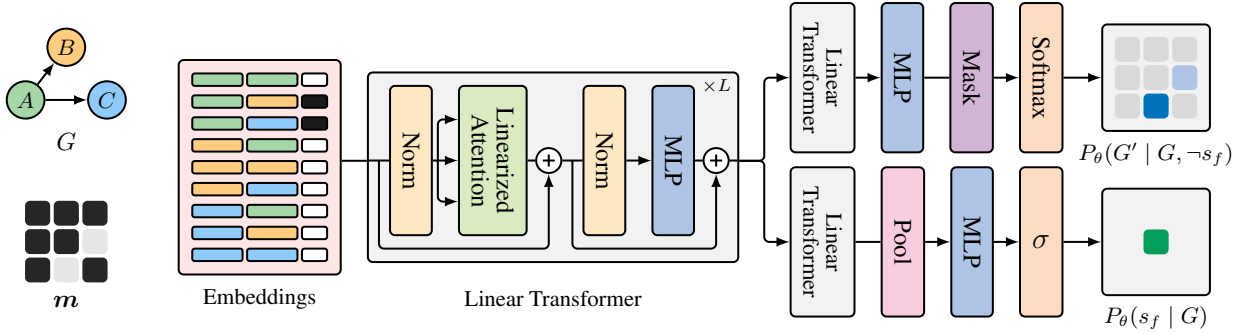
\begin{figure*}[t!]
    \centering
    \begin{tikzpicture}[every node/.style={inner sep=0pt}, var/.style={draw=black, minimum size=15pt, circle, font=\small, thick, anchor=center, fill=white}, factor/.style={fill=black, minimum size=5pt, rectangle, anchor=center}, edge/.style={thick, -latex}, modulation/.style={thick, -|}, zlabel/.style={font=\scriptsize, yshift=-2pt}, observed/.style={fill=gray!30}, flowlabel/.style={midway, font=\scriptsize}, rewardlabel/.style={pos=1, font=\scriptsize, right}, gfn_edge/.style={ultra thick, -latex}, gfn_node/.style={ultra thick, circle, draw, text height=3.5em, fill=gray!10}, gfn_other/.style={pos=1, anchor=west, inner sep=2pt}, adjacency/.style={minimum width=10pt, minimum height=10pt, rectangle, rounded corners=2pt, font=\bfseries\scriptsize}, block/.style={rounded corners=2pt, draw=black, thick, minimum height=1.5em, minimum width=6em, align=center, inner xsep=5pt, inner ysep=5pt}, transformer/.style={block, inner ysep=0pt}, plus/.style={draw=black, thick, fill=white, circle, minimum size=1em}, vector/.style={inner sep=0pt, minimum height=5pt, minimum width=2em, draw=black, thick, rounded corners=1pt}, mini vector/.style={vector, minimum width=1em}, y=12pt, x=50pt]

\definecolor{colorA}{HTML}{a5d6a7}  
\definecolor{colorB}{HTML}{ffcc80}  
\definecolor{colorC}{HTML}{90caf9}  

\node[transformer, fill=gray!10, rotate=270] (transformer) at (0, 0) {\begin{tikzpicture}
    \node[block, fill=Dandelion!30] (norm1) {Norm};
    \node[block, fill=LimeGreen!30, above=1em of norm1] (mha) {Linearized\\[-0.2em]Attention};
    
    \node[plus, above=3pt of mha] (plus1) {};
    \draw[thick, shorten <=3pt, shorten >=3pt] (plus1.north) -- (plus1.south);
    \draw[thick, shorten <=3pt, shorten >=3pt] (plus1.west) -- (plus1.east);
    
    \node[block, fill=Dandelion!30, above=0.8em of plus1] (norm2) {Norm};
    \node[block, fill=RoyalBlue!30, above=1em of norm2] (mlp) {MLP};
    
    \node[plus, above=3pt of mlp] (plus2) {};
    \draw[thick, shorten <=3pt, shorten >=3pt] (plus2.north) -- (plus2.south);
    \draw[thick, shorten <=3pt, shorten >=3pt] (plus2.west) -- (plus2.east);
    
    \coordinate (middle) at ($(norm1.north)!0.2!(mha.south)$);
    \draw[edge] (norm1) -- (mha);
    \draw[edge, rounded corners=2pt] (middle) -| (mha.220);
    \draw[edge, rounded corners=2pt] (middle) -| (mha.320);
    \draw[thick] (mha) -- (plus1);
    \draw[thick] (norm1.south) -- ++(270:0.8em) coordinate[midway] (bottom_norm1);
    \coordinate[right=5pt of mha] (right_mha);
    \draw[edge, rounded corners=2pt] (bottom_norm1) -| (right_mha) |- (plus1);
    \draw[thick] (plus1) -- (norm2) coordinate[midway] (bottom_norm2);
    \draw[edge] (norm2) -- (mlp);
    \draw[thick] (mlp) -- (plus2);
    \draw[thick] (plus2.north) -- ++(90:3pt);
    \coordinate[right=5pt of mlp] (right_mlp);
    \draw[edge, rounded corners=2pt] (bottom_norm2) -| (right_mlp) |- (plus2);
\end{tikzpicture}};

\node[block, fill=red!10, left=1em of transformer.south] (embedding) {\begin{tikzpicture}
    \node[vector, fill=colorA] (v11) {};
    \node[vector, fill=colorA, right=1pt of v11] (v12) {};
    \node[mini vector, fill=white, right=1pt of v12] (v13) {};
    
    \node[vector, fill=colorA, below=3pt of v11] (v21) {};
    \node[vector, fill=colorB, right=1pt of v21] (v22) {};
    \node[mini vector, fill=black!90, right=1pt of v22] (v23) {};
    
    \node[vector, fill=colorA, below=3pt of v21] (v31) {};
    \node[vector, fill=colorC, right=1pt of v31] (v32) {};
    \node[mini vector, fill=black!90, right=1pt of v32] (v33) {};
    
    \node[vector, fill=colorB, below=3pt of v31] (v41) {};
    \node[vector, fill=colorA, right=1pt of v41] (v42) {};
    \node[mini vector, fill=white, right=1pt of v42] (v43) {};
    
    \node[vector, fill=colorB, below=3pt of v41] (v51) {};
    \node[vector, fill=colorB, right=1pt of v51] (v52) {};
    \node[mini vector, fill=white, right=1pt of v52] (v53) {};
    
    \node[vector, fill=colorB, below=3pt of v51] (v61) {};
    \node[vector, fill=colorC, right=1pt of v61] (v62) {};
    \node[mini vector, fill=white, right=1pt of v62] (v63) {};
    
    \node[vector, fill=colorC, below=3pt of v61] (v71) {};
    \node[vector, fill=colorA, right=1pt of v71] (v72) {};
    \node[mini vector, fill=white, right=1pt of v72] (v73) {};
    
    \node[vector, fill=colorC, below=3pt of v71] (v81) {};
    \node[vector, fill=colorB, right=1pt of v81] (v82) {};
    \node[mini vector, fill=white, right=1pt of v82] (v83) {};
    
    \node[vector, fill=colorC, below=3pt of v81] (v91) {};
    \node[vector, fill=colorC, right=1pt of v91] (v92) {};
    \node[mini vector, fill=white, right=1pt of v92] (v93) {};
\end{tikzpicture}};
\draw[thick] ([yshift=2.6pt]embedding.east) -- ++(0:1em);

\node[left=2em of embedding.west] (graph_mask) {\begin{tikzpicture}
\node[] (graph) at (0, 0) {\begin{tikzpicture}
    \node[var, fill=colorA] (A) {$A$};
    \node[var, fill=colorB, above right=1em and 0.5em of A] (B) {$B$};
    \node[var, fill=colorC, below right=1em and 0.5em of B] (C) {$C$};
    
    \draw[edge] (A) -- (B);
    \draw[edge] (A) -- (C);
\end{tikzpicture}};
\node[below=0.5em of graph] (graph_label) {$G$};

\node[below=2em of graph_label] (mask) {\begin{tikzpicture}
    \node[adjacency, fill=black!85] (AA) {};
    \node[adjacency, fill=black!85, right=1pt of AA] (AB) {};
    \node[adjacency, fill=black!85, right=1pt of AB] (AC) {};
    
    \node[adjacency, fill=black!85, below=1pt of AA] (BA) {};
    \node[adjacency, fill=black!85, right=1pt of BA] (BB) {};
    \node[adjacency, fill=gray!20, right=1pt of BB] (BC) {};
    
    \node[adjacency, fill=black!85, below=1pt of BA] (CA) {};
    \node[adjacency, fill=gray!20, right=1pt of CA] (CB) {};
    \node[adjacency, fill=black!85, right=1pt of CB] (CC) {};
\end{tikzpicture}};
\node[below=0.5em of mask] (mask_label) {$\vm$};

\end{tikzpicture}};

\node[font=\scriptsize, anchor={north east}, xshift=-3pt, yshift=-3pt] (timesL) at (transformer.{north west}) {$\times L$};

\node[below=0.5em of embedding, font=\small, align=center] (embedding_label) {Embeddings};
\node[font=\small] (transformer_label) at (embedding_label -| transformer.east) {Linear Transformer};

\node[block, fill=gray!10, rotate=270, above right=3.6em and 2em of {transformer.north}, anchor=south, font=\small, align=center] (logits_head) {Linear\\[-0.2em]Transformer};
\node[block, fill=RoyalBlue!30, rotate=270, right=1em of logits_head.north, anchor=south] (logits_head_1) {MLP};
\node[block, fill=Plum!30, rotate=270, right=1em of logits_head_1.north, anchor=south] (logits_head_2) {Mask};
\coordinate (mid_logits) at ($(transformer.north)!0.3!(logits_head)$);
\draw[edge, rounded corners=2pt] ([yshift=2.6pt]transformer.north) -| (mid_logits) |- (logits_head.south);
\draw[edge] (logits_head) -- (logits_head_1);
\draw[thick] (logits_head_1) -- (logits_head_2);

\node[block, fill=gray!10, rotate=270, below right=3em and 2em of {transformer.north}, anchor=south, font=\small, align=center] (stop_head) {Linear\\[-0.2em]Transformer};
\node[block, fill=Rhodamine!30, rotate=270, right=1em of stop_head.north , anchor=south] (stop_head_1) {Pool};
\node[block, fill=RoyalBlue!30, rotate=270, right=1em of stop_head_1.north, anchor=south] (stop_head_2) {MLP};
\coordinate (mid_stop) at ($(transformer.north)!0.3!(stop_head)$);
\draw[edge, rounded corners=2pt] ([yshift=2.6pt]transformer.north) -| (mid_stop) |- (stop_head.south);
\draw[thick] (stop_head) -- (stop_head_1);
\draw[edge] (stop_head_1) -- (stop_head_2);

\node[block, fill=Peach!30, rotate=270, right=1em of logits_head_2.north, anchor=south] (softmax) {Softmax};
\draw[edge] (logits_head_2) -- (softmax);

\node[right=1.5em of softmax.north, inner sep=5pt, rounded corners=2pt, fill=gray!10, draw=black, thick] (adjacency_out) {\begin{tikzpicture}
    \node[adjacency, fill=gray!30] (AA) {};
    \node[adjacency, fill=gray!30, right=1pt of AA] (AB) {};
    \node[adjacency, fill=gray!30, right=1pt of AB] (AC) {};
    
    \node[adjacency, fill=gray!30, below=1pt of AA] (BA) {};
    \node[adjacency, fill=gray!30, right=1pt of BA] (BB) {};
    \node[adjacency, fill=white!70!RoyalBlue, right=1pt of BB] (BC) {};
    
    \node[adjacency, fill=gray!30, below=1pt of BA] (CA) {};
    \node[adjacency, fill=white!00!RoyalBlue, right=1pt of CA] (CB) {};
    \node[adjacency, fill=gray!30, right=1pt of CB] (CC) {};
\end{tikzpicture}};
\draw[edge] (softmax.north) -- (adjacency_out);

\node[block, fill=Peach!30, rotate=270, right=1em of stop_head_2.north, anchor=south] (sigmoid) {\vphantom{Softmax}};
\node[font=\large] (sigmoid_label) at (sigmoid.center) {$\sigma$};
\draw[edge] (stop_head_2) -- (sigmoid);

\node[right=1.5em of sigmoid.north, inner sep=5pt, rounded corners=2pt, fill=gray!10, draw=black, thick] (stop_out) {\begin{tikzpicture}
    \node[adjacency] (AA) {};
    \node[adjacency, right=1pt of AA] (AB) {};
    \node[adjacency, right=1pt of AB] (AC) {};
    
    \node[adjacency, below=1pt of AA] (BA) {};
    \node[adjacency, fill=white!5!ForestGreen, right=1pt of BA] (BB) {};
    \node[adjacency, right=1pt of BB] (BC) {};
    
    \node[adjacency, below=1pt of BA] (CA) {};
    \node[adjacency, right=1pt of CA] (CB) {};
    \node[adjacency, right=1pt of CB] (CC) {};
\end{tikzpicture}};
\draw[edge] (sigmoid.north) -- (stop_out);

\node[below=0em of adjacency_out, anchor=north, font=\small, inner sep=4pt] (logits_label) {$P_{\theta}(G'\mid G, \neg s_{f})$};
\node[below=0em of stop_out, anchor=north, font=\small, inner sep=4pt] (stop_label) {$P_{\theta}(s_{f}\mid G)$};

\end{tikzpicture}
    \vspace*{-0.5em}
    \caption{Neural network architecture of the forward transition probabilities $P_{\theta}(G_{t+1} \mid G_{t})$. The input graph $G$ is encoded as a set of $d^{2}$ possible edges (including self-loops). Each directed edge is embedded using the embeddings of its source and target, with an additional vector indicating whether the edge is present in $G$. These embeddings are fed into a Linear Transformer \citep{katharopoulos2020lineartransformers}, with two separate output heads. The first head (above) gives the probability to add a new edge $P_{\theta}(G' \mid G, \neg s_{f})$, using the mask $\vm$ associated to $G$ to filter out invalid actions; here, the only valid actions are either adding $B \rightarrow C$, or $C \rightarrow B$. The second head (below) gives the probability to terminate the trajectory $P_{\theta}(s_{f}\mid G)$.}
    \label{fig:transformer-architecture}
\end{figure*}
Beyond having enough capacity to satisfy as well as possible the detailed-balance condition at all states, we choose to parametrize the forward transition probabilities with neural networks to benefit from their capacity to generalize to states not encountered during training. In practice, instead of defining two separate networks to parametrize $P_{\theta}(s_{f} \mid G)$ and $P_{\theta}(G' \mid G, \neg s_{f})$, we use a single neural network with a common backbone and two separate heads, to benefit from parameter sharing. The full architecture is given in \cref{fig:transformer-architecture}.

Our choice of neural network architecture is motivated by multiple factors: we want an architecture (1) that is invariant to the order of the inputs, since $G$ is represented as a set of edges, (2) that transforms a set of input edges into a set of output probabilities for each edge to be added, in order to define $P_{\theta}(G'\mid G, \neg s_{f})$, and (3) whose parameters $\theta$ do not scale too much with $d$. A natural option would be to use a Transformer \citep{vaswani2017transformer}; however, because the size of our inputs is $d^{2}$, the self-attention layers would scale as $d^{4}$, and this would severely limit our ability to apply our method to model a distribution over larger DAGs.

We opted for a Linear Transformer \citep{katharopoulos2020lineartransformers} instead, which has the advantage to not suffer from this quadratic scaling in the input size. This architecture relies on a linearized attention mechanism, defined as
\begin{align}
    Q = \vx W_{Q} \qquad K = \vx W_{K} \qquad V = \vx W_{V}\nonumber\\
    \mathrm{LinAttn}_{k}(\vx) = \frac{\sum_{j=1}^{J}\big(\phi(Q_{k})^{\top}\phi(K_{j})\big)V_{j}}{\sum_{j=1}^{J}\phi(Q_{k})^{\top}\phi(K_{j})},
\end{align}
where $\vx$ is the input of the linearized attention layer, $\phi(\cdot)$ is a non-linear feature map, $J$ is the size of the input $\vx$ (in our case, $J = d^{2}$), and $Q$, $K$, and $V$ are linear transformations of $\vx$ corresponding to the queries, keys, and values respectively, as is standard with Transformers.

\section{Application to Bayesian Structure Learning}
\label{sec:application-bayesian-structure-learning}
We are given a dataset $\gD = \{\vx^{(1)}, \ldots, \vx^{(N)}\}$ of $N$ observations $\vx^{(j)}$, each consisting of $d$ elements. We consider the task of characterizing the posterior distribution $P(G \mid \gD)$ over Bayesian networks that model these observations. We assume that the samples in $\gD$ are iid. and fully-observed. As an alternative to MCMC \citep{madigan1995structuremcmc} or variational inference \citep{lorch2021dibs}, we approximate the posterior distribution over DAGs using a GFlowNet, as described in the previous section. For any DAG $G$, we will define its reward as the joint probability
\begin{equation}
    R(G) = P(G)P(\gD \mid G),
    \label{eq:gflownet-reward}
\end{equation}
where $P(G)$ is a prior over DAGs \citep{eggeling2019structureprior}, and $P(\gD \mid G)$ is the marginal likelihood. In Sec.~\ref{sec:detailed-balance-condition}, we saw that if the detailed-balance conditions are satisfied for all the states of the GFlowNet, then this yields a sampling process with probability proportional to $R(G)$. Therefore, by Bayes' theorem, a GFlowNet with the specific reward function in \cref{eq:gflownet-reward} approximates the posterior distribution $P(G\mid \gD) \propto R(G)$. We call our method \emph{DAG-GFlowNet}.

\subsection{Modularity \& computational efficiency}
\label{sec:modularity-computational-efficiency}
Following prior works on Bayesian structure learning, we assume that both the priors over parameters $P(\phi \mid G)$ of the Bayesian network (required to compute the marginal likelihood) and over structures $P(G)$ are \emph{modular} \citep{heckerman1995bde,chickering1995learning}. As a consequence the reward $R(G)$ is also modular, and its logarithm can be written as a sum of local scores that only depend on individual variables and their parents in $G$:
\begin{equation}
    \log R(G) = \sum_{j=1}^{d}\mathrm{LocalScore}\big(X_{j}\mid \Pa_{G}(X_{j})\big).
    \label{eq:log-reward-modular}
\end{equation}
Note that with our choice of reward, $\log R(G)$ corresponds to the Bayesian score \citep{koller2009pgm}. Examples of modular scores include the BDe score \citep{heckerman1995bde} and the BGe score \citep{geiger1994bge,kuipers2014bgeaddendum}. In order to fit the parameters $\theta$ of the GFlowNet, we will use the detailed-balance loss in \cref{eq:detailed-balance-loss}. We can observe that this loss function only involves the difference in log-rewards $\log R(G') - \log R(G)$ between two consecutive states, where $G'$ is the result of adding some edge $X_{i} \rightarrow X_{j}$ to the DAG $G$. Using our assumption of modularity, we can therefore compute this difference efficiently, as the terms in \cref{eq:log-reward-modular} remain unchanged for $j' \neq j$:
\begin{align}
    &\log R(G') - \log R(G) = \mathrm{LocalScore}\big(X_{j}\mid \Pa_{G}(X_{j}) \cup \{X_{i}\}\big)\nonumber\\
    &\qquad \qquad - \mathrm{LocalScore}\big(X_{j} \mid \Pa_{G}(X_{j})\big).
    \label{eq:delta-score}
\end{align}
This difference in local scores is sometimes called the \emph{delta score}, or the \emph{incremental value} \citep{friedman2003ordermcmc}, and has been employed in the literature to improve the efficiency of search algorithms \citep{chickering2002ges,koller2009pgm}.
\begin{figure*}[t!]
    \centering
    \begin{tikzpicture}[]

\pgfplotsset{every x tick label/.append style={font=\scriptsize, yshift=0ex}}
\pgfplotsset{every y tick label/.append style={font=\scriptsize, xshift=0ex}}

\begin{groupplot}[
    group style={group size=3 by 1, horizontal sep=3em}
]

\nextgroupplot[
    ymajorgrids,
    xmajorgrids,
    axis equal,
    ylabel={DAG-GFlowNet},
    xlabel={Exact posterior},
    ymin=0,
    ymax=1,
    xmin=0,
    xmax=1,
    height=6cm,
    width=6cm,
    legend style={font=\small},
    minor x tick num=1,
    minor y tick num=1,
]

\addplot[color=black, thick, dashed, mark=none] coordinates {(0, 0) (1, 1)};

\addplot[color=BrickRed, mark=*, only marks, mark size=1.2pt, opacity=0.5] table[col sep=comma, x=full, y=estimate] {figures/small-graphs/edge.csv};

\nextgroupplot[
    ymajorgrids,
    xmajorgrids,
    axis equal,
    xlabel={Exact posterior},
    ymin=0,
    ymax=1,
    xmin=0,
    xmax=1,
    height=6cm,
    width=6cm,
    legend style={font=\small},
    minor x tick num=1,
    minor y tick num=1,
]

\addplot[color=black, thick, dashed, mark=none] coordinates {(0, 0) (1, 1)};

\addplot[color=BrickRed, mark=*, only marks, mark size=1.2pt, opacity=0.5] table[col sep=comma, x=full, y=estimate] {figures/small-graphs/path.csv};

\nextgroupplot[
    ymajorgrids,
    xmajorgrids,
    axis equal,
    xlabel={Exact posterior},
    ymin=0,
    ymax=1,
    xmin=0,
    xmax=1,
    height=6cm,
    width=6cm,
    legend style={font=\small},
    minor x tick num=1,
    minor y tick num=1,
]

\addplot[color=black, thick, dashed, mark=none] coordinates {(0, 0) (1, 1)};

\addplot[color=BrickRed, mark=*, only marks, mark size=1.2pt, opacity=0.5] table[col sep=comma, x=full, y=estimate] {figures/small-graphs/markov_blanket.csv};

\end{groupplot}

\node[below=3em of {group c1r1}.south] (edge) {(a) Edge features};
\node[below=3em of {group c2r1}.south] (path) {(b) Path features};
\node[below=3em of {group c3r1}.south] (markov) {(c) Markov features};

\node[anchor=south east, above left=3pt of {group c1r1}.{south east}, inner sep=3pt, fill=white] (r_edge) {$r = 0.9992$};
\node[anchor=south east, above left=3pt of {group c2r1}.{south east}, inner sep=3pt, fill=white] (r_path) {$r = 0.9989$};
\node[anchor=south east, above left=3pt of {group c3r1}.{south east}, inner sep=3pt, fill=white] (r_markov) {$r = 0.9997$};

\end{tikzpicture}
    \caption{Comparison between the exact posterior distribution and the posterior approximation from DAG-GFlowNet, for different structural features: (a) edge features $X_{i} \rightarrow X_{j}$, (b) path features $X_{i} \rightsquigarrow X_{j}$, (c) Markov features $X_{i} \sim_{M} X_{j}$. Each point corresponds to a feature computed for specific variables $X_{i}$ and $X_{j}$ in a graph over $d = 5$ nodes, either based on the exact posterior (x-axis), or the posterior approximation found with the GFlowNet (y-axis). We repeated this experiment with $20$ different (ground-truth) DAGs. The Pearson correlation coefficient $r$ is included in the bottom-right corner of each~plot.}
    \label{fig:posterior-comparison}
\end{figure*}
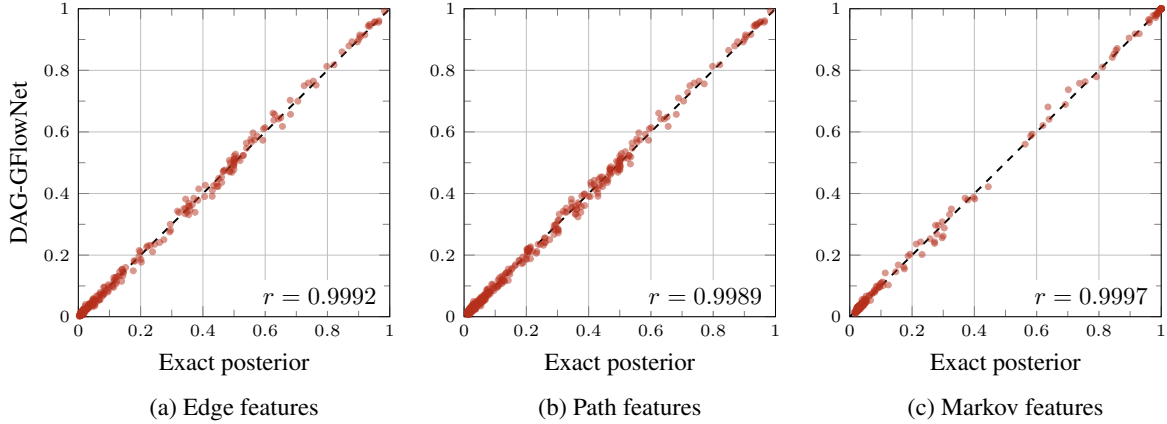

\subsection{Off-policy learning}
\label{sec:off-policy-learning}
As the number of states in DAG-GFlowNet is super-exponential in $d$, the number of nodes in each DAG $G$, it would be impractical to minimize the detailed-balance loss for all possible transitions $G \rightarrow G'$. Alternatively, we can minimize this loss in expectation using a distribution $\pi(G \rightarrow G')$ with full support over transitions:
\begin{equation}
    \gL(\theta) = \E_{\pi}\Bigg[\bigg[\log \frac{R(G')P_{B}(G \mid G')P_{\theta}(s_{f}\mid G)}{R(G)P_{\theta}(G'\mid G)P_{\theta}(s_{f}\mid G')}\bigg]^{2}\Bigg].
    \label{eq:expected-detailed-balance-loss}
\end{equation}
This distribution $\pi(G \rightarrow G')$ can be arbitrary; for example, we can use $P_{\theta}(G' \mid G)$ directly and learn it \emph{on-policy} \citep{rummery1994sarsa}, as long as it assigns non-zero probability to any next state $G'$.

Taking inspiration from Deep Q-learning \citep{mnih2015dqn}, we instead learn $P_{\theta}$ using \emph{off-policy} data. Transitions $G \rightarrow G'$ are collected based on $P_{\theta}(G'\mid G)$, along with their corresponding delta score \cref{eq:delta-score}, and they are stored in a replay buffer. We can also sample some transitions uniformly at random, with probability $\varepsilon$, to encourage exploration. To estimate $\gL(\theta)$ and update the parameters $\theta$, we can then sample a mini-batch of transitions randomly from the replay buffer. Moreover, again inspired by Deep Q-learning \citep{van2018deadlytriad}, we found it advantageous to evaluate $P_{\bar{\theta}}(s_{f} \mid G')$ in \cref{eq:expected-detailed-balance-loss} with a separate target network---where the parameters $\bar{\theta}$ are updated periodically.

\section{Experimental results}
\label{sec:experimental-reults}
We compared DAG-GFlowNet against 3 broad classes of Bayesian structure learning algorithms: MCMC, non-parametric DAG Bootstrapping \citep{friedman1999bootstrap}, and variational inference. We used Structure MCMC \citep[MC\textsuperscript{3};][]{madigan1995structuremcmc} and the recent Gadget \citep{viinikka2020gadget} samplers as two representative methods based on MCMC. Following \citet{lorch2021dibs}, we used two variants of Bootstrapping based on the score-based algorithm GES \citep[Bootstrap GES;][]{chickering2002ges}, and the constraint-based algorithm PC \citep[Bootstrap PC;][]{spirtes2000pc}, as the internal structure learning routines. Finally for methods based on variational inference, we used DiBS \citep{lorch2021dibs} and BCD Nets \citep{cundy2021bcdnets}. Throughout this section, we used the BGe score for continuous data, and the BDe score for discrete data, to compute $\log p(\gD \mid G)$.

\subsection{Comparison with the exact posterior}
\label{sec:comparison-exact-posterior}
\begin{figure*}[t!]
    \centering
    \begin{tikzpicture}[]

\pgfplotsset{every y tick label/.append style={font=\scriptsize, xshift=0ex}}
\pgfplotsset{every x tick label/.append style={font=\scriptsize, yshift=0ex}}

\definecolor{color_mc3}{RGB}{114,158,206}
\definecolor{color_gadget}{RGB}{174,199,232}
\definecolor{color_bpc}{RGB}{255,158,74}
\definecolor{color_bges}{RGB}{255,187,120}
\definecolor{color_dibs}{RGB}{103,191,92}
\definecolor{color_bcdnets}{RGB}{152,223,138}
\definecolor{color_gflownet}{RGB}{237,102,93}

\begin{groupplot}[
    group style={group size=3 by 1, horizontal sep=3em}
]


\nextgroupplot[
    ymajorgrids,
    xmajorgrids,
    title={$\mathbb{E}$-SHD},
    xlabel={},
    height=6cm,
    width=8cm,
    legend style={font=\small},
    scaled x ticks=false,
    scaled y ticks=false,
    boxplot/draw direction=y,
    enlarge x limits=0.02,
    x tick label style={rotate=0},
    ytick={20, 40, 60, 80, 100},
    minor y tick num=1,
    enlarge x limits=0.1,
    xmin=1, xmax=7,
    xtick={1, 2, 3, 4, 5, 6, 7},
    xticklabels={
        MC\textsuperscript{3},
        Gadget\vphantom{\textsuperscript{3}},
        B-PC\vphantom{\textsuperscript{3}},
        B-GES\vphantom{\textsuperscript{3}},
        DiBS\vphantom{\textsuperscript{3}},
        BCD\vphantom{\textsuperscript{3}},
        GFN\vphantom{\textsuperscript{3}}
    },
]

\addplot+ [boxplot prepared={
    lower whisker=48.84,
    lower quartile=54.19,
    median=58.75,
    upper quartile=63.56,
    upper whisker=71.30
},
thick, solid, mark=x, black, fill=color_mc3,
mark options={fill=black, draw=black}] table [row sep=\\, y index=0] { \\ };

\addplot+ [boxplot prepared={
    lower whisker=73.80,
    lower quartile=78.26,
    median=80.02,
    upper quartile=82.49,
    upper whisker=87.78
},
thick, solid, mark=x, black, fill=color_gadget,
mark options={fill=black, draw=black}] table [row sep=\\, y index=0] { 90.50\\ 94.95\\ };

\addplot+ [boxplot prepared={
    lower whisker=23.59,
    lower quartile=29.04,
    median=33.18,
    upper quartile=38.28,
    upper whisker=48.67
},
thick, solid, mark=x, black, fill=color_bpc,
mark options={fill=black, draw=black}] table [row sep=\\, y index=0] { \\ };

\addplot+ [boxplot prepared={
    lower whisker=36.03,
    lower quartile=39.68,
    median=43.03,
    upper quartile=51.37,
    upper whisker=68.04
},
thick, solid, mark=x, black, fill=color_bges,
mark options={fill=black, draw=black}] table [row sep=\\, y index=0] { \\ };

\addplot+ [boxplot prepared={
    lower whisker=32.69,
    lower quartile=41.02,
    median=42.76,
    upper quartile=48.74,
    upper whisker=57.68
},
thick, solid, mark=x, black, fill=color_dibs,
mark options={fill=black, draw=black}] table [row sep=\\, y index=0] { 62.84\\ };

\addplot+ [boxplot prepared={
    lower whisker=13.73,
    lower quartile=19.99,
    median=23.36,
    upper quartile=28.62,
    upper whisker=39.25
},
thick, solid, mark=x, black, fill=color_bcdnets,
mark options={fill=black, draw=black}] table [row sep=\\, y index=0] { \\ };

\addplot+ [boxplot prepared={
    lower whisker=25.16,
    lower quartile=28.21,
    median=32.50,
    upper quartile=41.09,
    upper whisker=49.74
},
thick, solid, mark=x, black, fill=color_gflownet,
mark options={fill=black, draw=black}] table [row sep=\\, y index=0] { \\ };


\nextgroupplot[
    ymajorgrids,
    xmajorgrids,
    title={AUROC},
    xlabel={},
    height=6cm,
    width=8cm,
    legend style={font=\small},
    scaled x ticks=false,
    scaled y ticks=false,
    boxplot/draw direction=y,
    enlarge x limits=0.02,
    x tick label style={rotate=0},
    enlarge x limits=0.1,
    minor y tick num=1,
    xmin=1, xmax=7,
    xtick={1, 2, 3, 4, 5, 6, 7},
    xticklabels={
        MC\textsuperscript{3},
        Gadget\vphantom{\textsuperscript{3}},
        B-PC\vphantom{\textsuperscript{3}},
        B-GES\vphantom{\textsuperscript{3}},
        DiBS\vphantom{\textsuperscript{3}},
        BCD\vphantom{\textsuperscript{3}},
        GFN\vphantom{\textsuperscript{3}}
    },
]

\addplot+ [boxplot prepared={
    lower whisker=0.669,
    lower quartile=0.771,
    median=0.801,
    upper quartile=0.847,
    upper whisker=0.921
},
thick, solid, mark=x, black, fill=color_mc3,
mark options={fill=black, draw=black}] table [row sep=\\, y index=0] { \\ };

\addplot+ [boxplot prepared={
    lower whisker=0.386,
    lower quartile=0.481,
    median=0.523,
    upper quartile=0.564,
    upper whisker=0.667
},
thick, solid, mark=x, black, fill=color_gadget,
mark options={fill=black, draw=black}] table [row sep=\\, y index=0] { \\ };

\addplot+ [boxplot prepared={
    lower whisker=0.626,
    lower quartile=0.725,
    median=0.749,
    upper quartile=0.793,
    upper whisker=0.855
},
thick, solid, mark=x, black, fill=color_bpc,
mark options={fill=black, draw=black}] table [row sep=\\, y index=0] { \\ };

\addplot+ [boxplot prepared={
    lower whisker=0.803,
    lower quartile=0.869,
    median=0.894,
    upper quartile=0.926,
    upper whisker=0.973
},
thick, solid, mark=x, black, fill=color_bges,
mark options={fill=black, draw=black}] table [row sep=\\, y index=0] { 0.709\\ 0.780\\ };

\addplot+ [boxplot prepared={
    lower whisker=0.762,
    lower quartile=0.811,
    median=0.830,
    upper quartile=0.851,
    upper whisker=0.905
},
thick, solid, mark=x, black, fill=color_dibs,
mark options={fill=black, draw=black}] table [row sep=\\, y index=0] { \\ };

\addplot+ [boxplot prepared={
    lower whisker=0.693,
    lower quartile=0.753,
    median=0.779,
    upper quartile=0.805,
    upper whisker=0.864
},
thick, solid, mark=x, black, fill=color_bcdnets,
mark options={fill=black, draw=black}] table [row sep=\\, y index=0] { \\ };

\addplot+ [boxplot prepared={
    lower whisker=0.661,
    lower quartile=0.778,
    median=0.813,
    upper quartile=0.868,
    upper whisker=0.928
},
thick, solid, mark=x, black, fill=color_gflownet,
mark options={fill=black, draw=black}] table [row sep=\\, y index=0] { \\ };


\nextgroupplot[
    ymajorgrids,
    xmajorgrids,
    title style={yshift=-2pt},
    title={$\log P(G, \mathcal{D}' \mid \mathcal{D})$},
    xlabel={},
    height=6cm,
    width=8cm,
    legend style={font=\small},
    scaled x ticks=false,
    scaled y ticks=false,
    boxplot/draw direction=y,
    x tick label style={rotate=0},
    xtick={1, 2, 3, 4, 5, 6, 7},
    ytick={700, 750, 800, 850, 900},
    minor y tick num=4,
    enlarge x limits=0.1,
    xmin=1, xmax=7,
    xticklabels={
        MC\textsuperscript{3},
        Gadget\vphantom{\textsuperscript{3}},
        B-PC\vphantom{\textsuperscript{3}},
        B-GES\vphantom{\textsuperscript{3}},
        DiBS\vphantom{\textsuperscript{3}},
        BCD\vphantom{\textsuperscript{3}},
        GFN\vphantom{\textsuperscript{3}}
    },
]

\addplot[mark=none, gray, thick, densely dashed, samples=2, domain=0:8] {904.21};

\addplot+ [boxplot prepared={
    lower whisker=861.23,
    lower quartile=873.07,
    median=878.20,
    upper quartile=882.05,
    upper whisker=892.93
},
thick, solid, mark=x, black, fill=color_mc3,
mark options={fill=black, draw=black}] table [row sep=\\, y index=0] { \\ };

\addplot+ [boxplot prepared={
    lower whisker=752.80,
    lower quartile=793.71,
    median=807.13,
    upper quartile=821.28,
    upper whisker=862.55
},
thick, solid, mark=x, black, fill=color_gadget,
mark options={fill=black, draw=black}] table [row sep=\\, y index=0] { 745.02\\ 748.45\\ 752.03\\ 745.83\\ };

\addplot+ [boxplot prepared={
    lower whisker=725.87,
    lower quartile=769.57,
    median=786.84,
    upper quartile=799.73,
    upper whisker=843.79
},
thick, solid, mark=x, black, fill=color_bpc,
mark options={fill=black, draw=black}] table [row sep=\\, y index=0] { 720.89\\ 709.87\\ 721.50\\ 720.64\\ 850.27\\ };

\addplot+ [boxplot prepared={
    lower whisker=864.54,
    lower quartile=881.60,
    median=888.14,
    upper quartile=893.20,
    upper whisker=907.31
},
thick, solid, mark=x, black, fill=color_bges,
mark options={fill=black, draw=black}] table [row sep=\\, y index=0] { 860.45\\ 861.07\\ 858.10\\ 863.71\\ 863.90\\ 862.81\\ 855.84\\ 863.77\\ 863.43\\ 860.73\\ 858.15\\ };

\addplot+ [boxplot prepared={
    lower whisker=831.80,
    lower quartile=862.80,
    median=874.44,
    upper quartile=883.91,
    upper whisker=907.57
},
thick, solid, mark=x, black, fill=color_dibs,
mark options={fill=black, draw=black}] table [row sep=\\, y index=0] { 823.53\\ 818.40\\ 826.60\\ 830.04\\ 828.63\\ 816.10\\ 824.51\\ 828.40\\ 815.25\\ 829.42\\ 830.22\\ 808.24\\ 791.84\\ 824.19\\ 825.62\\ 828.90\\ 818.87\\ 828.55\\ 816.54\\ 826.19\\ 822.69\\ 822.13\\ };

\addplot+ [boxplot prepared={
    lower whisker=875.83,
    lower quartile=884.54,
    median=889.10,
    upper quartile=891.53,
    upper whisker=898.48
},
thick, solid, mark=x, black, fill=color_bcdnets,
mark options={fill=black, draw=black}] table [row sep=\\, y index=0] { 868.09\\ 872.25\\ 865.23\\ 867.74\\ 867.62\\ 864.31\\ 864.36\\ 871.42\\ 858.65\\ 873.33\\ };

\addplot+ [boxplot prepared={
    lower whisker=874.95,
    lower quartile=887.47,
    median=892.21,
    upper quartile=896.09,
    upper whisker=905.68
},
thick, solid, mark=x, black, fill=color_gflownet,
mark options={fill=black, draw=black}] table [row sep=\\, y index=0] { 864.32\\ 871.60\\ 867.75\\ 867.71\\ 871.09\\ 872.72\\ 868.05\\ };

\end{groupplot}

\end{tikzpicture}
    \caption{Bayesian structure learning of linear-Gaussian Bayesian networks with $d = 20$ nodes. Results for $\E$-SHD \& AUROC are aggregated over 25 randomly generated datasets $\gD$, sampled from different (ground-truth) Bayesian networks. Results for $\log P(G, \gD' \mid \gD)$ are given for a single dataset $\gD$; the dashed line corresponds to the log-likelihood of the ground truth graph $G^{\star}$. For $\E$-SHD lower is better, and for AUROC and $\log P(G, \gD' \mid \gD)$ higher is better. Labels: B-PC = Bootstrap-PC, B-GES = Bootstrap-GES, BCD = BCD Nets, GFN = DAG-GFlowNet.}
    \label{fig:lingauss20}
\end{figure*}
In order to measure the quality of the posterior approximation returned by DAG-GFlowNet, we want to compare it with the exact posterior distribution $P(G\mid \gD)$. However, the latter requires an exhaustive enumeration of all possible DAGs, which is only feasible for graphs with no more than $5$ nodes. Therefore, we sampled $N=100$ datapoints from a randomly generated (under an Erd\H{o}s-R\'{e}nyi model; \citealp{erdos1960ergraphs}) linear-Gaussian Bayesian network over $d=5$ variables. We used the BGe score to compute the reward $R(G) = P(G)P(\gD \mid G)$. The exact posterior distribution $P(G\mid \gD)$ is obtained by enumerating all $29,\!281$ possible DAGs over $5$ nodes and computing their respective rewards $R(G)$ (normalized to sum to $1$).

We evaluated the quality of the approximation based on the probability of various structural features. For example, using samples $\{G_{1}, G_{2}, \ldots, G_{n}\}$ from the posterior approximation, the marginal probability of an \emph{edge feature} $X_{i} \rightarrow X_{j}$ can be estimated with
\begin{equation}
    P_{\theta}(X_{i} \rightarrow X_{j}\mid \gD) \approx \frac{1}{n}\sum_{k=1}^{n} \mathbf{1}(X_{i} \rightarrow X_{j} \in G_{k}),
    \label{eq:edge-features}
\end{equation}
where $\mathbf{1}(\cdot)$ is the indicator function. For the exact posterior, we can obtain the posterior probability of the edge feature by simply marginalizing over $P(G\mid \gD)$. Similarly, we compute (or estimate) the marginal probability of a \emph{path feature} ${X_{i} \rightsquigarrow X_{j}}$, i.e. of a (directed) path existing from $X_{i}$ to $X_{j}$, and the probability of a \emph{Markov feature} ${X_{i} \sim_{M} X_{j}}$, i.e. of $X_{i}$ being in the Markov blanket of $X_{j}$ \citep{friedman2003ordermcmc}. These features are computed for all variables $X_{i}$ and $X_{j}$ in the Bayesian network.

In \cref{fig:posterior-comparison}, we compare the probabilities of these features for both the exact posterior and the distribution induced by DAG-GFlowNet, where we repeated the experiment above with $20$ different (ground-truth) Bayesian networks. We observe that the probabilities of all structural features estimated by the GFlowNet are strongly correlated with the exact marginal probabilities. This shows that DAG-GFlowNet is capable of learning a very accurate approximation of the posterior distribution over graphs $P(G\mid \gD)$.

\subsection{Simulated data}
\label{sec:simulated-data}
We follow the experimental setup of \citet{zheng2018notears} \& \citet{lorch2021dibs}, and sample synthetic data from linear-Gaussian Bayesian networks with randomly generated structures; details about this data generation process are given in \cref{app:simulated-data-lingauss50}. To show that DAG-GFlowNet can effectively approximate the posterior distribution over larger graphs, we experimented with Bayesian networks of size $d = 20$ (and $d = 50$, see \cref{app:simulated-data-lingauss50}). Similar to \cref{sec:comparison-exact-posterior}, the ground-truth graphs are sampled according to an Erd\H{o}s-R\'{e}nyi model, with $2d$ edges in expectation---a setting sometimes referred to as ER2 \citep{cundy2021bcdnets}. For each experiment, we sampled a dataset $\gD$ of $N = 100$ observations, and we used the BGe score to compute $R(G)$.

Since we have access to the ground-truth graph $G^{\star}$ that generated $\gD$, we evaluate the performance of each algorithm with the \emph{expected structural Hamming distance} ($\E$-SHD) to $G^{\star}$ over the posterior approximation; a detailed definition is available in \cref{app:details-metrics}. We also compute the \emph{area under the ROC curve} \citep[AUROC;][]{husmeier2003auroc} for the edge features as defined in \cref{eq:edge-features}, compared to the edges of $G^{\star}$. Finally, we compute the joint log-likelihood $\log P(G, \gD'\mid \gD)$ on a held-out dataset $\gD'$; we chose this metric over the log-predictive likelihood $\log P(\gD'\mid \gD)$, as proposed by \citet{eaton2007bayesian}, to study the effect of the posterior approximation $P(G \mid \gD)$.

The results on graphs with $d = 20$ nodes are shown in \cref{fig:lingauss20}. We observe that both in terms of $\E$-SHD \& AUROC, DAG-GFlowNet, is competitive against all other methods, in particular those based on MCMC, and this does not come at a cost in terms of its predictive capacity on held-out data. In particular, we can see that the distribution induced by DAG-GFlowNet yields a predictive log-likelihood concentrated near the log-likelihood of the ground-truth DAG $G^{\star}$.

\subsection{Application: Flow Cytometry Data}
\label{sec:application-flow-cytometry-data}
We also evaluated DAG-GFlowNet on real-world flow cytometry data \citep{sachs2005causal} to learn protein signaling pathways. The data consists of continuous measurements of $d = 11$ phosphoproteins in individual T-cells. Out of all the measurements, we selected the $N = 853$ observations corresponding to the first experimental condition of \citet{sachs2005causal} as our dataset $\gD$. Following prior work on structure learning, we used the DAG inferred by \citet{sachs2005causal}, containing $d = 11$ nodes and $17$ edges, as our graph of reference (ground-truth). However, it should be noted that this ``consensus graph'' may not represent a realistic and complete description of the system being modeled here \citep{mooij2020jci}. We standardized the data, and used the BGe score to compute $R(G)$.

\begin{table}[ht]
    \centering
    \caption{Learning protein signaling pathways from flow cytometry data \citep{sachs2005causal}. All results include a $95\%$ confidence interval estimated with bootstrap resampling.}
    \label{tab:sachs-continuous}
    \begin{adjustbox}{center, scale=0.9}
    \begin{tabular}{lccc}
        \toprule
         & $\E$-\#~Edges & $\E$-SHD  & AUROC \\
        \midrule
        MC\textsuperscript{3} & $10.96 \pm 0.09$ & $22.66 \pm 0.11$ & $0.508$ \\
        Gadget & $10.59 \pm 0.09$ & $21.77 \pm 0.10$ & $0.479$ \\
        Bootstrap GES & $11.11 \pm 0.09$ & $23.07 \pm 0.11$ & $\mathbf{0.548}$ \\
        Bootstrap PC & $\phantom{1}7.83 \pm 0.04$ & $20.65 \pm 0.06$ & $0.520$ \\
        DiBS & $12.62 \pm 0.16$ & $23.32 \pm 0.14$ & $0.518$ \\
        BCD Nets & $\phantom{1}4.14 \pm 0.09$ & $\mathbf{18.14 \pm 0.09}$ & $0.510$ \\
        \midrule
        DAG-GFlowNet & $11.25 \pm 0.09$ & $22.88 \pm 0.10$ & $0.541$ \\
        \bottomrule
    \end{tabular}
    \end{adjustbox}
\end{table}

In \cref{tab:sachs-continuous}, we compare the expected SHD and the AUROC obtained with DAG-GFlowNet and other approaches. While BCD Nets and Bootstrap PC have a smaller $\E$-SHD, suggesting that the distribution is concentrated closer to the consensus graph, in reality they tend to be more conservative and sample graphs with fewer edges. Overall, DAG-GFlowNet offers a good trade-off between performance (as measured by the $\E$-SHD and the AUROC), and getting a distribution that assigns higher probability to DAGs with more edges. We also observed that $1.50\%$ of the graphs sampled with DiBS contained a cycle.

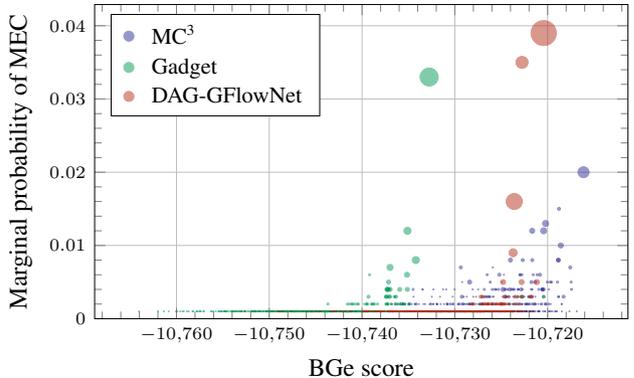
\begin{figure}[t!]
    \centering
    \begin{tikzpicture}[]

\pgfplotsset{every x tick label/.append style={font=\scriptsize, yshift=0ex}}
\pgfplotsset{every y tick label/.append style={font=\scriptsize, xshift=0ex}}
\pgfplotsset{
    jitter/.style={
        y filter/.code={\pgfmathparse{\pgfmathresult+0.5*rand*#1}}
    },
    jitter/.default=0.1
}
\pgfplotsset{tick scale binop=\times}

\begin{axis}[
    ymajorgrids,
    xmajorgrids,
    ylabel={Marginal probability of MEC},
    xlabel={BGe score},
    height=6cm,
    width=9cm,
    legend style={font=\small},
    scaled x ticks=false,
    scaled y ticks=false,
    yticklabel style = {
       /pgf/number format/fixed,
       /pgf/number format/precision=2,
    },
    legend pos=north west,
    legend cell align={left},
    legend style={column sep=5pt, inner xsep=5pt},
    minor x tick num=4,
    minor y tick num=4,
    enlarge y limits=upper,
    ymin=0
]

\addplot[
    scatter=true,
    mark=*,
    only marks,
    opacity=0.5,
    point meta=explicit symbolic,
    scatter/@pre marker code/.style={/tikz/mark size=\pgfplotspointmeta/5},
    scatter/@post marker code/.style={},
    color=Blue,
] table[col sep=comma, x=score, y=marginal, meta=unique_dags] {figures/sachs-comparison/mc3.csv};
\addlegendentry{MC\textsuperscript{3}}

\addplot[
    scatter=true,
    mark=*,
    only marks,
    opacity=0.5,
    point meta=explicit symbolic,
    scatter/@pre marker code/.style={/tikz/mark size=\pgfplotspointmeta/5},
    scatter/@post marker code/.style={},
    color=ForestGreen,
] table[col sep=comma, x=score, y=marginal, meta=unique_dags] {figures/sachs-comparison/gadget.csv};
\addlegendentry{Gadget}

\addplot[
    scatter=true,
    mark=*,
    only marks,
    opacity=0.5,
    point meta=explicit symbolic,
    scatter/@pre marker code/.style={/tikz/mark size=\pgfplotspointmeta/5},
    scatter/@post marker code/.style={},
    color=BrickRed,
] table[col sep=comma, x=score, y=marginal, meta=unique_dags] {figures/sachs-comparison/gflownet.csv};
\addlegendentry{DAG-GFlowNet}

\end{axis}

\end{tikzpicture}
    \caption{Coverage of the posterior approximations learned on flow cytometry data \citep{sachs2005causal}. Each point corresponds to a sampled Markov equivalence class, and its size represents the number of different DAGs (in the equivalence class) sampled from the posterior approximation. See \cref{fig:sachs-comparison-all} in \cref{app:sachs-comparison} for an additional comparison with methods based on Variational Inference.}
    \label{fig:sachs-comparison-mcmc}
\end{figure}

Beyond these metrics, we would like to test if the advantages of Bayesian structure learning are also reflected in the distribution induced by DAG-GFlowNet. In particular, we want to study (1) if this distribution covers multiple high-scoring DAGs, instead of being peaked at a single most likely graph, and (2) if the GFlowNet can sample a variety of DAGs from the same Markov equivalence class (MEC), showing the inherent uncertainty over equivalent graphs. In \cref{fig:sachs-comparison-mcmc}, we visualize the MECs of the graphs sampled with DAG-GFlowNet, and two methods based on MCMC (MC\textsuperscript{3} and Gadget); other baselines were excluded for clarity. The size of each point represents the number of unique DAGs in the corresponding MEC. We observe that DAG-GFlowNet largely follows the behavior of MCMC: the distribution does not collapse to a single most-likely DAG, and covers multiple MECs. Moreover, the GFlowNet is also capable of sampling different equivalent DAGs (corresponding to larger points), showing again that the distribution does not collapse to a single representative of the MECs with higher marginal probability. We also observe that the maximum a posteriori MEC found by DAG-GFlowNet reaches a higher score than the one found with Gadget, but a lower score than MC\textsuperscript{3}; as a point of reference, the score of the best MEC obtained with GES \citep{chickering2002ges} is $-10,\!716.12$.

\subsection{Application: Interventional data}
\label{sec:application-interventional-data}
In addition to the observational data we used in \cref{sec:application-flow-cytometry-data}, \citet{sachs2005causal} also provided flow cytometry data under different experimental conditions, where the T-cells were perturbed with some reagents; this effectively corresponds to interventional data \citep{pearl2009causality}. Although a molecular intervention may be imperfect and affect multiple proteins \citep{eaton2007belief}, we assume here that these interventions are perfect, and the intervention targets are known. We used a discretized dataset of $N = 5,\!400$ samples from $9$ experimental conditions---of which $6$ are interventions. We modified the BDe score to handle this mixture of observational and interventional data \citep{cooper1999interventional}.
\begin{table}[ht]
    \centering
    \caption{Combining discrete interventional and observational flow cytometry data \citep{sachs2005causal}. ${}^{\star}$Result reported in~\citet{eaton2007belief}.}
    \label{tab:sachs-interventional}
    \begin{adjustbox}{center, scale=0.9}
    \begin{tabular}{lccc}
        \toprule
         & $\E$-\#~Edges & $\E$-SHD  & AUROC \\
        \midrule
        Exact posterior${}^{\star}$ & --- & --- & $\mathbf{0.816}$ \\
        MC\textsuperscript{3} & $25.97 \pm 0.01$ & $\mathbf{25.08 \pm 0.02}$ & $0.665$ \\
        \midrule
        DAG-GFlowNet & $30.66 \pm 0.04$ & $27.77 \pm 0.03$ & $0.700$ \\
        \bottomrule
    \end{tabular}
    \end{adjustbox}
\end{table}

In \cref{tab:sachs-interventional}, we compare with \citet{eaton2007belief}, which compute the AUROC of the exact posterior using dynamic programming, therefore working as an upper bound for what a posterior approximation can achieve. They achieve this at the expense of computing only edge marginals, without providing access to a distribution over DAGs. We also use the modified BDe score with MC\textsuperscript{3}, which predicts sparser graphs with higher SHD than DAG-GFlowNet, but lower AUROC. Note that this setup is different from previous works which use continuous data instead \citep{brouillard2020differentiable,faria2022differentiable}.

\section{Conclusion}
\label{sec:conclusion}
We have proposed a new method for Bayesian structure learning, based on a novel class of probabilistic models called GFlowNets, where the generation of a sample graph is treated as a sequential decision problem. We introduced a number of enhancements to the standard framework of GFlowNets, specifically designed for approximating a distribution over DAGs. In cases where the data is limited and measuring the epistemic uncertainty is critical, DAG-GFlowNet offers an effective solution to approximate the posterior distribution over DAGs $P(G \mid \gD)$. However, we also observed that in its current state, DAG-GFlowNet may suffer from some limitations, notably as the size of the dataset $\gD$ increases; see \cref{app:limitations} for a discussion.

While DAG-GFlownet operates on the space of DAGs directly, the structure of the GFlowNet may eventually be adapted to work with alternative representations of statistical dependencies in Bayesian networks, such as essential graphs for MECs \citep{chickering2002ges}. Moreover, although we have already shown that DAG-GFlowNet can approximate the posterior using a mixture of observational and interventional data, we will continue to study in future work its applications to causal discovery, especially in the context of learning the structure of models with latent variables.

\begin{acknowledgements}
We would like to thank Emmanuel Bengio, Paul Bertin, and Valentin Thomas for the useful discussions about the project, and Dinghuai Zhang, Kolya Malkin, and Xu Ji for their valuable feedback on the paper.
This research was partially supported by the Canada CIFAR AI Chair Program and by Samsung Electronics Co., Ldt. Simon Lacoste-Julien is a CIFAR Associate Fellow in the Learning in Machines \& Brains program, Yoshua Bengio is a CIFAR Senior Fellow and Stefan Bauer is a CIFAR Azrieli Global Scholar.
\end{acknowledgements}

\bibliography{references}

\title{Bayesian Structure Learning with Generative Flow Networks\newline(Supplementary material)}

\makeatletter
\renewcommand{\AB@affillist}{}
\renewcommand{\AB@authlist}{}
\setcounter{authors}{0}
\renewcommand{\@thanks}{}
\makeatother

\author[1]{Tristan~Deleu}
\author[1]{Ant\'{o}nio~G\'{o}is}
\author[2]{Chris~Emezue}
\author[1]{Mansi~Rankawat}
\author[1,4]{\authorcr{Simon~Lacoste-Julien}}
\author[3,5]{Stefan~Bauer}
\author[1,4,6]{Yoshua~Bengio}
\affil[1]{Mila, Universit\'{e} de Montr\'{e}al}
\affil[2]{Technical University of Munich}
\affil[3]{KTH Stockholm\protect\\[0.5em]}
\affil[4]{CIFAR AI Chair}
\affil[5]{CIFAR Azrieli Global Scholar}
\affil[6]{CIFAR Senior Fellow}

\onecolumn
\maketitle
\appendix
\renewcommand{\theequation}{\thesection.\arabic{equation}}
\setcounter{equation}{0}
\setcounter{figure}{5}

\section{Limitations of DAG-GFlowNet}
\label{app:limitations}
Although we have shown in the main paper that DAG-GFlowNet is capable of learning an accurate approximation of the posterior distribution $P(G\mid \gD)$ when the size of the dataset $\gD$ is moderate (a situation where the benefits of a Bayesian treatment of structure learning are larger), we observed that as the size of the dataset increases, fitting the detailed-balance loss in \cref{eq:expected-detailed-balance-loss} was more challenging. This can be explained by the fact that with a larger amount of data, the posterior distribution becomes very peaky \citep{koller2009pgm}. As a consequence, in this situation, the delta-score in \cref{eq:delta-score}, which is required to calculate the loss, can take a wide range of values: adding an edge to a graph can drastically increase or decrease its score. In turn, the neural network parametrizing $P_{\theta}(G_{t+1}\mid G_{t})$ needs to compensate for these large fluctuations, making it harder to train.

Unfortunately, some of the standard techniques used in Machine Learning to tackle this issue, such as normalization of the inputs, cannot be applied here. Normalizing the delta-score is equivalent to normalizing the rewards $R(G)$ and $R(G')$ themselves, and as a consequence it would change the distribution that is being approximated: instead of approximating the posterior distribution $P(G \mid \gD)$, we would approximate a distribution $P(G\mid \gD)^{\tau}$ under some temperature $\tau$. Solutions to this problem include a schedule of temperature, similar to simulated annealing, or a reparametrization of $P_{\theta}(G_{t+1}\mid G_{t})$ to better handle large fluctuations of delta-scores; this exploration is left as future work.

\section{Detailed-balance condition with all complete states}
\label{app:detailed-balance-condition}
In this section, we will prove a special case of the \emph{detailed-balance condition} introduced by \citet{bengio2021gflownetfoundations} applied to the case where all the states of the GFlowNet are complete (except the terminal state $s_{f}$). To simplify the presentation, we will follow the notations of \citet{bengio2021gflownetfoundations}, and denote the forward transition probability by $P_{F}(s_{t+1} \mid s_{t})$---instead of $P_{\theta}(s_{t+1} \mid s_{t})$ in the main paper. Recall that the detailed-balance condition \citep[][Def. 17]{bengio2021gflownetfoundations} is given by
\begin{equation}
    F(s_{t})P_{F}(s_{t+1}\mid s_{t}) = F(s_{t+1})P_{B}(s_{t}\mid s_{t+1}).
    \label{eq:vanilla-detailed-balance-condition}
\end{equation}
In the case where all the states are complete, we also know that \citep[][Def. 16]{bengio2021gflownetfoundations}
\begin{equation*}
    P_{F}(s_{f}\mid s_{t}) := \frac{F(s_{t} \rightarrow s_{f})}{\sum_{s'\in\mathrm{Ch}(s_{t})}F(s_{t} \rightarrow s')} = \frac{R(s_{t})}{F(s_{t})} \quad \Leftrightarrow \quad F(s_{t}) = \frac{R(s_{t})}{P_{F}(s_{f}\mid s_{t})},
\end{equation*}
where $F(s \rightarrow s')$ represents the flow from state $s$ to $s'$, as described in \cref{sec:generative-flow-networks}, $F(s)$ is the total flow through state $s$, and we used Proposition~4 \& Equation~34 of \citet{bengio2021gflownetfoundations} to introduce $F(s_{t})$ and $R(s_{t})$ respectively. Replacing $F(\cdot)$ in \cref{eq:vanilla-detailed-balance-condition} yields the expected condition:
\begin{equation}
    R(s_{t})P_{F}(s_{t+1}\mid s_{t})P_{F}(s_{f}\mid s_{t+1}) = R(s_{t+1})P_{B}(s_{t}\mid s_{t+1})P_{F}(s_{f}\mid s_{t}).
\end{equation}

The original formulation in \cref{eq:vanilla-detailed-balance-condition} would require us to parametrize both $P_{F}(s_{t+1} \mid s_{t})$ and $F(s)$. On the other hand, using this alternative condition, we only have to parametrize $P_{F}(s_{t+1} \mid s_{t})$ (including when $s_{t+1} = s_{f}$ is the terminal state).

\section{Definition and update of the mask over actions}
\label{app:mask}
In \cref{sec:structure-gflownet}, we introduced a mask $\vm$ associated with a DAG $G$ to indicate which edges could be legally added to $G$ to obtain a new valid DAG $G'$. This mask must ignore (1) the edges already present in $G$ (which cannot be added further), and (2) any edge whose addition leads to the introduction of a cycle. The mask $\vm$ is constructed using (1) the adjacency matrix of $G$, and (2) the adjacency matrix of the transitive closure of $G^{\top}$, the transpose graph of $G$; recall that $G^{\top}$ is obtained from $G$ by inverting the direction of its edges.

\citet{giudici2003improvingmcmc} use a similar construction to efficiently obtain the legal actions their MCMC sampler may take. In particular, they show that this mask $\vm$ can be updated very efficiently online as edges are added one by one. In practice, this allows us to circumvent an expensive check for cycles at every stage of the construction of a sample DAG in the GFlowNet. Since the mask can be composed in 2 parts (as explained above), we can simply update each part anytime a new edge is added to a DAG $G$.

\begin{figure*}[t]
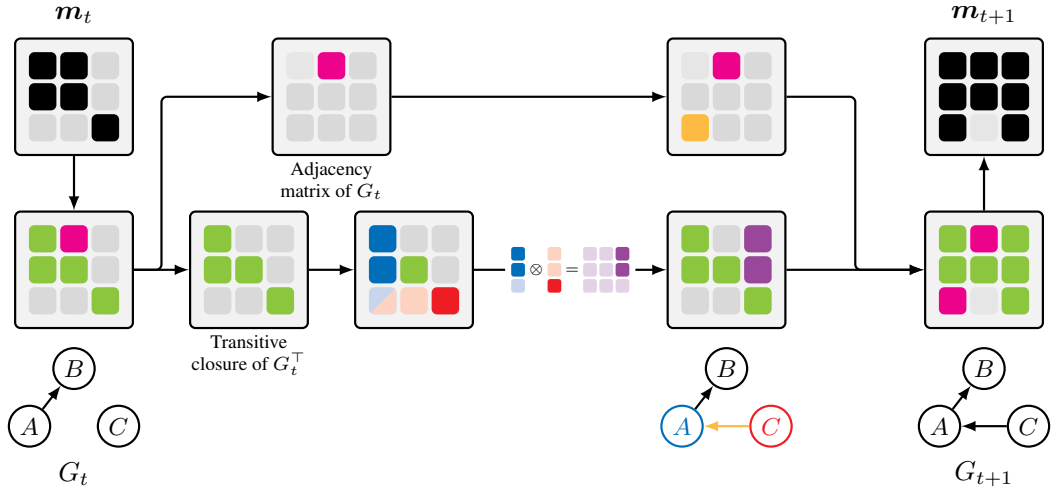

    \centering
    \includestandalone[width=0.8\linewidth]{figures/online-update-mask}
    \caption{Online update of the mask $\vm$. The mask $\vm_{t}$ associated with $G_{t}$ represents (in black) the edges that cannot be added to $G_{t}$ to obtain a valid DAG. $\vm_{t}$ is decomposed in two parts: the adjacency matrix of $G_{t}$ (top), and the transitive closure of $G_{t}^{\top}$ (bottom). To update the mask and obtain $\vm_{t+1}$ associated with $G_{t+1}$, the result of adding the edge $C \rightarrow A$ to $G_{t}$, each component must be updated separately, and then recombined. The diagonal elements of $\vm_{t}$, corresponding to self-loops (which are always invalid actions to take) are integrated into the transitive closure of $G_{t}^{\top}$ by convention.}
    \label{fig:online-update-mask}
\end{figure*}

In \cref{fig:online-update-mask}, we show how the mask $\vm_{t}$ associated with a graph $G_{t}$ can be updated after adding a new edge $C \rightarrow A$ to obtain the mask $\vm_{t+1}$. The mask is decomposed in 2 parts: the adjacency matrix of $G_{t}$, and the transitive closure of $G_{t}^{\top}$. After adding $C \rightarrow A$, each component is updated separately:
\begin{enumerate}
    \item \textbf{Adjacency matrix:} To update the adjacency matrix, the entry in the adjacency matrix must be set (here, the entry corresponding to the edge $C \rightarrow A$).
    \item \textbf{Transitive closure:} To update the transitive closure of the transpose, we need to compute the outer product of the column corresponding to the target of the edge (here $A$, in blue) with the row corresponding to the source of the edge (here $C$, in red). The outer product is added (more precisely, this is a binary OR) to the initial transitive closure.
\end{enumerate}
These two operations can be done very efficiently in $O(d^{2})$, where $d$ is the number of nodes in the DAG.

\section{Additional experimental results}
\label{app:additional-experimental-results}

\subsection{Details about the metrics}
\label{app:details-metrics}
Throughout this paper, we used mainly two metrics to compare the performance of DAG-GFlowNet over alternative Bayesian structure learning algorithms: the \emph{expected SHD} ($\E\mathrm{-SHD}$), and the \emph{area under the ROC curve} (AUROC). Let $\{G_{1}, \ldots, G_{n}\}$ be samples from the posterior approximation to be evaluated, and $G^{\star}$ be the ground truth graph. The $\E$-SHD to $G^{\star}$ can be estimated as
\begin{equation}
    \E\mathrm{-SHD} \approx \frac{1}{n}\sum_{k=1}^{n}\mathrm{SHD}(G_{k}, G^{\star}),
\end{equation}
where $\mathrm{SHD}(G, G^{\star})$ counts the number of edges changes (adding, removing, reversing an edge) necessary to move from $G$ to $G^{\star}$.

\subsection{Simulated data}
\label{app:simulated-data-lingauss50}
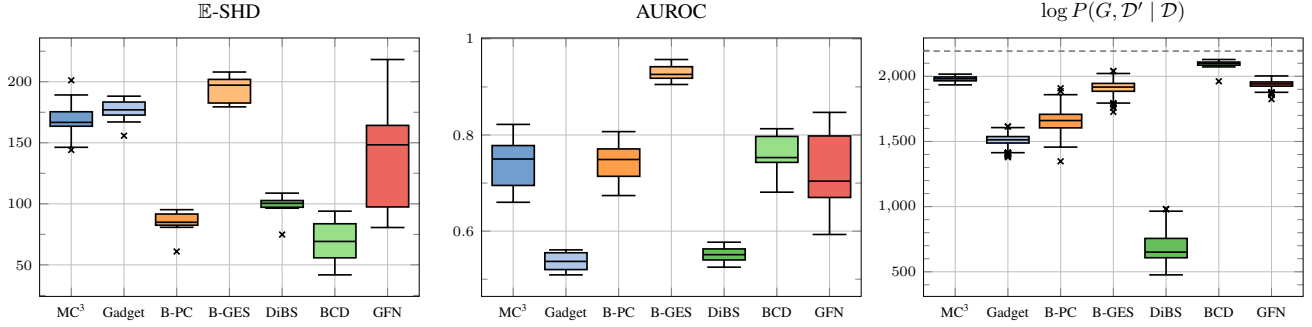
\begin{figure*}[t]
    \centering
    \begin{tikzpicture}[]

\pgfplotsset{every y tick label/.append style={font=\scriptsize, xshift=0ex}}
\pgfplotsset{every x tick label/.append style={font=\scriptsize, yshift=0ex}}

\definecolor{color_mc3}{RGB}{114,158,206}
\definecolor{color_gadget}{RGB}{174,199,232}
\definecolor{color_bpc}{RGB}{255,158,74}
\definecolor{color_bges}{RGB}{255,187,120}
\definecolor{color_dibs}{RGB}{103,191,92}
\definecolor{color_bcdnets}{RGB}{152,223,138}
\definecolor{color_gflownet}{RGB}{237,102,93}

\begin{groupplot}[
    group style={group size=3 by 1, horizontal sep=3em}
]


\nextgroupplot[
    ymajorgrids,
    xmajorgrids,
    title={$\mathbb{E}$-SHD},
    xlabel={},
    height=6cm,
    width=8cm,
    legend style={font=\small},
    scaled x ticks=false,
    scaled y ticks=false,
    boxplot/draw direction=y,
    enlarge x limits=0.02,
    x tick label style={rotate=0},
    minor y tick num=1,
    enlarge x limits=0.1,
    xmin=1, xmax=7,
    xtick={1, 2, 3, 4, 5, 6, 7},
    xticklabels={
        MC\textsuperscript{3},
        Gadget\vphantom{\textsuperscript{3}},
        B-PC\vphantom{\textsuperscript{3}},
        B-GES\vphantom{\textsuperscript{3}},
        DiBS\vphantom{\textsuperscript{3}},
        BCD\vphantom{\textsuperscript{3}},
        GFN\vphantom{\textsuperscript{3}}
    },
]

\addplot+ [boxplot prepared={
    lower whisker=146.29,
    lower quartile=163.55,
    median=166.71,
    upper quartile=175.41,
    upper whisker=189.17
},
thick, solid, mark=x, black, fill=color_mc3,
mark options={fill=black, draw=black}] table [row sep=\\, y index=0] { 201.16\\ 144.09\\ };

\addplot+ [boxplot prepared={
    lower whisker=167.05,
    lower quartile=172.70,
    median=176.96,
    upper quartile=183.37,
    upper whisker=188.17
},
thick, solid, mark=x, black, fill=color_gadget,
mark options={fill=black, draw=black}] table [row sep=\\, y index=0] { 155.80\\ };

\addplot+ [boxplot prepared={
    lower whisker=80.76,
    lower quartile=82.46,
    median=84.89,
    upper quartile=91.64,
    upper whisker=95.20
},
thick, solid, mark=x, black, fill=color_bpc,
mark options={fill=black, draw=black}] table [row sep=\\, y index=0] { 60.95\\ };

\addplot+ [boxplot prepared={
    lower whisker=179.42,
    lower quartile=182.56,
    median=197.20,
    upper quartile=201.90,
    upper whisker=207.95
},
thick, solid, mark=x, black, fill=color_bges,
mark options={fill=black, draw=black}] table [row sep=\\, y index=0] { \\ };

\addplot+ [boxplot prepared={
    lower whisker=96.42,
    lower quartile=97.18,
    median=100.39,
    upper quartile=102.71,
    upper whisker=108.76
},
thick, solid, mark=x, black, fill=color_dibs,
mark options={fill=black, draw=black}] table [row sep=\\, y index=0] { 74.83\\ };

\addplot+ [boxplot prepared={
    lower whisker=41.80,
    lower quartile=55.77,
    median=69.19,
    upper quartile=83.66,
    upper whisker=94.00
},
thick, solid, mark=x, black, fill=color_bcdnets,
mark options={fill=black, draw=black}] table [row sep=\\, y index=0] { \\ };

\addplot+ [boxplot prepared={
    lower whisker=80.60,
    lower quartile=97.44,
    median=148.34,
    upper quartile=164.23,
    upper whisker=218.21
},
thick, solid, mark=x, black, fill=color_gflownet,
mark options={fill=black, draw=black}] table [row sep=\\, y index=0] { \\ };


\nextgroupplot[
    ymajorgrids,
    xmajorgrids,
    title={AUROC},
    xlabel={},
    height=6cm,
    width=8cm,
    legend style={font=\small},
    scaled x ticks=false,
    scaled y ticks=false,
    boxplot/draw direction=y,
    enlarge x limits=0.02,
    x tick label style={rotate=0},
    enlarge x limits=0.1,
    minor y tick num=1,
    xmin=1, xmax=7,
    xtick={1, 2, 3, 4, 5, 6, 7},
    xticklabels={
        MC\textsuperscript{3},
        Gadget\vphantom{\textsuperscript{3}},
        B-PC\vphantom{\textsuperscript{3}},
        B-GES\vphantom{\textsuperscript{3}},
        DiBS\vphantom{\textsuperscript{3}},
        BCD\vphantom{\textsuperscript{3}},
        GFN\vphantom{\textsuperscript{3}}
    },
]

\addplot+ [boxplot prepared={
    lower whisker=0.660,
    lower quartile=0.695,
    median=0.750,
    upper quartile=0.778,
    upper whisker=0.822
},
thick, solid, mark=x, black, fill=color_mc3,
mark options={fill=black, draw=black}] table [row sep=\\, y index=0] { \\ };

\addplot+ [boxplot prepared={
    lower whisker=0.509,
    lower quartile=0.520,
    median=0.537,
    upper quartile=0.555,
    upper whisker=0.561
},
thick, solid, mark=x, black, fill=color_gadget,
mark options={fill=black, draw=black}] table [row sep=\\, y index=0] { \\ };

\addplot+ [boxplot prepared={
    lower whisker=0.674,
    lower quartile=0.714,
    median=0.749,
    upper quartile=0.771,
    upper whisker=0.807
},
thick, solid, mark=x, black, fill=color_bpc,
mark options={fill=black, draw=black}] table [row sep=\\, y index=0] { \\ };

\addplot+ [boxplot prepared={
    lower whisker=0.905,
    lower quartile=0.918,
    median=0.926,
    upper quartile=0.942,
    upper whisker=0.957
},
thick, solid, mark=x, black, fill=color_bges,
mark options={fill=black, draw=black}] table [row sep=\\, y index=0] { \\ };

\addplot+ [boxplot prepared={
    lower whisker=0.525,
    lower quartile=0.540,
    median=0.551,
    upper quartile=0.563,
    upper whisker=0.577
},
thick, solid, mark=x, black, fill=color_dibs,
mark options={fill=black, draw=black}] table [row sep=\\, y index=0] { \\ };

\addplot+ [boxplot prepared={
    lower whisker=0.681,
    lower quartile=0.743,
    median=0.753,
    upper quartile=0.797,
    upper whisker=0.813
},
thick, solid, mark=x, black, fill=color_bcdnets,
mark options={fill=black, draw=black}] table [row sep=\\, y index=0] { \\ };

\addplot+ [boxplot prepared={
    lower whisker=0.593,
    lower quartile=0.670,
    median=0.704,
    upper quartile=0.798,
    upper whisker=0.847
},
thick, solid, mark=x, black, fill=color_gflownet,
mark options={fill=black, draw=black}] table [row sep=\\, y index=0] { \\ };


\nextgroupplot[
    ymajorgrids,
    xmajorgrids,
    title style={yshift=-2pt},
    title={$\log P(G, \mathcal{D}' \mid \mathcal{D})$},
    xlabel={},
    height=6cm,
    width=8cm,
    legend style={font=\small},
    scaled x ticks=false,
    scaled y ticks=false,
    boxplot/draw direction=y,
    x tick label style={rotate=0},
    xtick={1, 2, 3, 4, 5, 6, 7},
    minor y tick num=4,
    enlarge x limits=0.1,
    xmin=1, xmax=7,
    xticklabels={
        MC\textsuperscript{3},
        Gadget\vphantom{\textsuperscript{3}},
        B-PC\vphantom{\textsuperscript{3}},
        B-GES\vphantom{\textsuperscript{3}},
        DiBS\vphantom{\textsuperscript{3}},
        BCD\vphantom{\textsuperscript{3}},
        GFN\vphantom{\textsuperscript{3}}
    },
]

gflownet

\addplot[mark=none, gray, thick, densely dashed, samples=2, domain=0:8] {2193.16};

\addplot+ [boxplot prepared={
    lower whisker=1934.19,
    lower quartile=1965.15,
    median=1981.20,
    upper quartile=1994.65,
    upper whisker=2016.55
},
thick, solid, mark=x, black, fill=color_mc3,
mark options={fill=black, draw=black}] table [row sep=\\, y index=0] { \\ };

\addplot+ [boxplot prepared={
    lower whisker=1413.86,
    lower quartile=1487.87,
    median=1512.71,
    upper quartile=1537.39,
    upper whisker=1606.23
},
thick, solid, mark=x, black, fill=color_gadget,
mark options={fill=black, draw=black}] table [row sep=\\, y index=0] { 1394.90\\ 1378.50\\ 1413.37\\ 1392.24\\ 1403.23\\ 1410.25\\ 1393.81\\ 1614.12\\ 1399.34\\ 1410.62\\ 1413.06\\ 1404.08\\ 1384.40\\ 1404.75\\ 1615.05\\ 1413.57\\ 1413.22\\ };

\addplot+ [boxplot prepared={
    lower whisker=1456.68,
    lower quartile=1604.15,
    median=1660.43,
    upper quartile=1707.97,
    upper whisker=1858.77
},
thick, solid, mark=x, black, fill=color_bpc,
mark options={fill=black, draw=black}] table [row sep=\\, y index=0] { 1347.51\\ 1908.33\\ 1885.49\\ };

\addplot+ [boxplot prepared={
    lower whisker=1794.46,
    lower quartile=1884.95,
    median=1917.24,
    upper quartile=1945.30,
    upper whisker=2021.39
},
thick, solid, mark=x, black, fill=color_bges,
mark options={fill=black, draw=black}] table [row sep=\\, y index=0] { 2039.98\\ 2040.16\\ 1726.83\\ 1789.30\\ 1793.91\\ 1763.53\\ 1758.07\\ 1782.82\\ 1785.83\\ };

\addplot+ [boxplot prepared={
    lower whisker=476.35,
    lower quartile=607.75,
    median=651.01,
    upper quartile=756.06,
    upper whisker=964.88
},
thick, solid, mark=x, black, fill=color_dibs,
mark options={fill=black, draw=black}] table [row sep=\\, y index=0] { 982.24\\ 979.21\\ };

\addplot+ [boxplot prepared={
    lower whisker=2071.86,
    lower quartile=2085.18,
    median=2098.85,
    upper quartile=2109.07,
    upper whisker=2128.67
},
thick, solid, mark=x, black, fill=color_bcdnets,
mark options={fill=black, draw=black}] table [row sep=\\, y index=0] { 1961.40\\ };

\addplot+ [boxplot prepared={
    lower whisker=1876.98,
    lower quartile=1924.32,
    median=1941.08,
    upper quartile=1956.49,
    upper whisker=2002.61
},
thick, solid, mark=x, black, fill=color_gflownet,
mark options={fill=black, draw=black}] table [row sep=\\, y index=0] { 1824.41\\ 1874.94\\ 1867.71\\ 1863.43\\ 1864.00\\ 1872.93\\ 1869.34\\ 1867.63\\ 1856.97\\ 1862.78\\ 1874.79\\ 1876.00\\ 1873.10\\ };

\end{groupplot}

\end{tikzpicture}
    \caption{Bayesian structure learning of linear-Gaussian Bayesian networks with $d = 50$ nodes. Results for $\E$-SHD \& AUROC are aggregated over 10 randomly generated datasets $\gD$, sampled from different (ground-truth) Bayesian networks. Results for $\log P(G, \gD' \mid \gD)$ are given for a single dataset $\gD$; the dashed line corresponds to the log-likelihood of the ground truth graph. Labels: B-PC = Bootstrap-PC, B-GES = Bootstrap-GES, BCD = BCD Nets, GFN = DAG-GFlowNet.}
    \label{fig:lingauss50}
\end{figure*}

In addition to the experiments on simulated data with graphs over $d=20$ nodes, we also compared DAG-GFlowNet with other methods on graphs with $d=50$ nodes. The experimental setup described in \cref{sec:simulated-data} remains unchanged, and the data generation process is detailed below. We show this comparison in \cref{fig:lingauss50}, in terms of $\E$-SHD, AUROC, and the joint log-likelihood $P(\gD', G \mid \gD)$ on some held-out dataset $\gD'$. We observe that DAG-GFlowNet is still competitive compared to the other algorithms, even though it suffers from a higher variance. This can be partly explained by the neural network parametrizing the forward transition probability $P_{\theta}(G_{t+1} \mid G_{t})$ (see \cref{sec:forward-transition-probabilities,sec:parametrization-linear-transformers}) underfitting the data, and therefore not accurately matching the detailed-balance conditions, necessary for a close approximation of the posterior distribution $P(G\mid \gD)$. Similar to our observations in \cref{sec:application-flow-cytometry-data}, we also noticed that algorithms that tend to perform better in terms of $\E$-SHD (e.g. BCD Nets, Bootstrap-PC) tend to have an order of magnitude fewer edges in the sampled DAGs.

\paragraph{Data generation} For our experiments on simulated data, we followed the generation process described in \citep{lorch2021dibs}. The data was generated in the following way:
\begin{enumerate}
    \item We sampled a DAG using an Erd\H{o}s-R\'{e}nyi model \citep{erdos1960ergraphs}, with $2d$ edges on average; the value of the probability of creating an edge between two nodes was scaled accordingly.
    \item Once the structure of the graph is known, we sampled the parameters of the linear-Gaussian model randomly from a standard Normal distribution $\gN(0, 1)$. The linear-Gaussian model is therefore defined as, $\forall j \in [1, d]$
    \begin{equation*}
        X_{j} = \sum_{X_{i} \in \mathrm{Pa}_{G}(X_{j})} \beta_{ij}X_{i} + \varepsilon,
    \end{equation*}
    where $\beta_{ij} \sim \gN(0, 1)$, and $\varepsilon \sim \gN(0, 0.01)$. The defines all the conditional probability distribution of the generative model.
    \item Once the full Bayesian Network is known, we used ancestral sampling to generate $N = 100$ datapoints to fill our dataset $\gD$.
\end{enumerate}

\subsection{Flow Cytometry Data}
\label{app:sachs-comparison}
In \cref{sec:application-flow-cytometry-data}, we described an application of DAG-GFlowNet to real-world flow cytometry data. In particular, we showed in \cref{fig:sachs-comparison-mcmc} that DAG-GFlowNet was capable of modeling a distribution that was not only capable of capturing the mode of the posterior distribution (i.e., graphs with a high score), but also had diversity in the graphs sampled, both in terms of the different Markov Equivalence Classes (MECs) those graphs belong to, but also multiple unique DAG instances of the same MEC (depicted by the size of each point in \cref{fig:sachs-comparison-mcmc}).

\begin{figure}[t!]
    \centering
    \begin{tikzpicture}[]

\pgfplotsset{every x tick label/.append style={font=\scriptsize, yshift=0ex}}
\pgfplotsset{every y tick label/.append style={font=\scriptsize, xshift=0ex}}
\pgfplotsset{
    jitter/.style={
        y filter/.code={\pgfmathparse{\pgfmathresult+0.5*rand*#1}}
    },
    jitter/.default=0.1
}
\pgfplotsset{tick scale binop=\times}

\begin{axis}[
    ymajorgrids,
    xmajorgrids,
    ylabel={log-marginal probability of MEC},
    xlabel={BGe score},
    height=8cm,
    width=18cm,
    legend style={font=\small},
    scaled x ticks=false,
    scaled y ticks=false,
    yticklabel style = {
       /pgf/number format/fixed,
       /pgf/number format/precision=2,
    },
    legend pos=north west,
    legend cell align={left},
    legend style={column sep=5pt, inner xsep=5pt},
    minor x tick num=4,
    minor y tick num=4,
    enlarge y limits,
    ymode=log
]

\addplot[
    scatter=true,
    mark=*,
    only marks,
    opacity=0.5,
    point meta=explicit symbolic,
    scatter/@pre marker code/.style={/tikz/mark size=\pgfplotspointmeta/5},
    scatter/@post marker code/.style={},
    color=Blue,
] table[col sep=comma, x=score, y=marginal, meta=unique_dags] {figures/sachs-comparison/mc3.csv};
\addlegendentry{MC\textsuperscript{3}}

\addplot[
    scatter=true,
    mark=*,
    only marks,
    opacity=0.5,
    point meta=explicit symbolic,
    scatter/@pre marker code/.style={/tikz/mark size=\pgfplotspointmeta/5},
    scatter/@post marker code/.style={},
    color=ForestGreen,
] table[col sep=comma, x=score, y=marginal, meta=unique_dags] {figures/sachs-comparison/gadget.csv};
\addlegendentry{Gadget}

\addplot[
    scatter=true,
    mark=*,
    only marks,
    opacity=0.5,
    point meta=explicit symbolic,
    scatter/@pre marker code/.style={/tikz/mark size=\pgfplotspointmeta/5},
    scatter/@post marker code/.style={},
    color=YellowOrange,
] table[col sep=comma, x=score, y=marginal, meta=unique_dags] {figures/sachs-comparison/dibs.csv};
\addlegendentry{DiBS}

\addplot[
    scatter=true,
    mark=*,
    only marks,
    opacity=0.5,
    point meta=explicit symbolic,
    scatter/@pre marker code/.style={/tikz/mark size=\pgfplotspointmeta/5},
    scatter/@post marker code/.style={},
    color=Purple,
] table[col sep=comma, x=score, y=marginal, meta=unique_dags] {figures/sachs-comparison/bcdnets.csv};
\addlegendentry{BCD Nets}

\addplot[
    scatter=true,
    mark=*,
    only marks,
    opacity=0.5,
    point meta=explicit symbolic,
    scatter/@pre marker code/.style={/tikz/mark size=\pgfplotspointmeta/5},
    scatter/@post marker code/.style={},
    color=BrickRed,
] table[col sep=comma, x=score, y=marginal, meta=unique_dags] {figures/sachs-comparison/gflownet.csv};
\addlegendentry{DAG-GFlowNet}

\end{axis}

\end{tikzpicture}
    \caption{Coverage of the posterior approximations learned on flow cytometry data \citep{sachs2005causal}. Each point corresponds to a sampled Markov equivalence class, and its size represents the number of different DAGs (in the equivalence class) sampled from the posterior approximation.}
    \label{fig:sachs-comparison-all}
\end{figure}
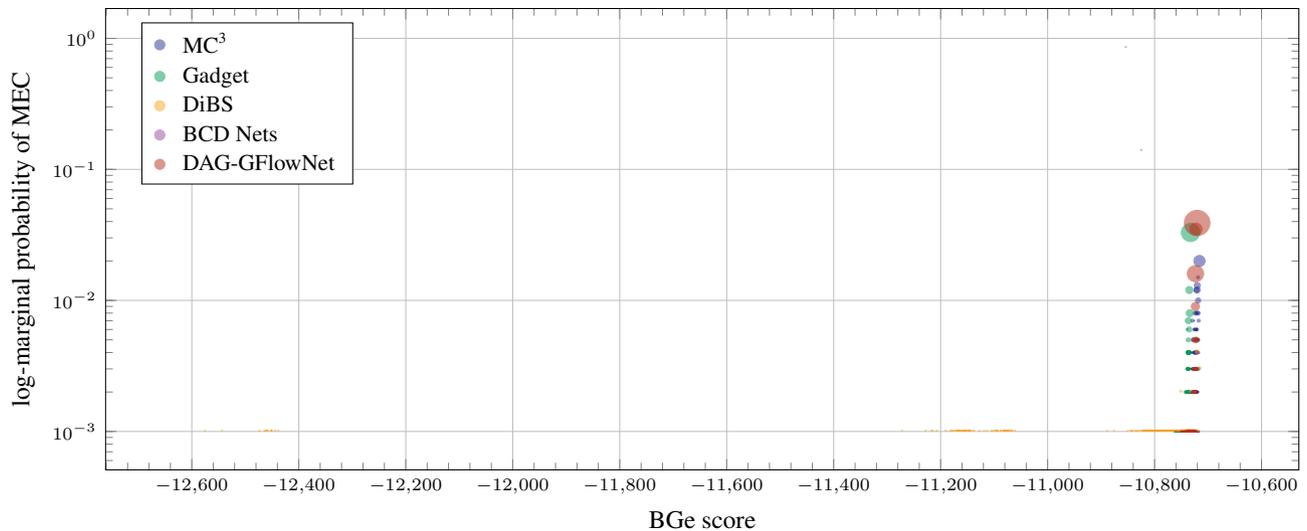

However for clarity, we only compared DAG-GFlowNet to methods based on MCMC in \cref{fig:sachs-comparison-mcmc}. In \cref{fig:sachs-comparison-all}, we also added a comparison to BCD Nets \citep{cundy2021bcdnets} and DiBS \citep{lorch2021dibs}, two methods based on Variational Inference. Although we saw in \cref{tab:sachs-continuous} that those two methods were comparing favorably against other algorithms in terms of $\E\mathrm{-SHD}$ and AUROC, including against DAG-GFlowNet, we can assess more precisely the quality of the posterior approximation returned by BCD Nets and DiBS:
\begin{itemize}
    \item Out of $1,\!000$ graphs sampled with BCD Nets, those graphs belonged to one of only two MECs (with a BGe score around $-10,\!950$). Furthermore, as shown by the size of each point, those MECs happen to only contain a single unique DAG. Overall, this means that BCD Nets only returned 2 unique DAGs (out of the $1,\!000$ samples), showing the lack of diversity of the posterior approximation learned with BCD Nets.
    \item DiBS sampled a significant number of very low scoring DAGs, with BGe scores as low as $-12,\!600$ (whereas the best MEC obtained with GES \citep{chickering2002ges} had a score of $-10,\!716.12$).
    \item With our choice of the BGe score, the true posterior distribution would assign the same probability to graphs in the same MEC. However, we can see that DiBS only returned graphs belonging to unique MECs, as opposed to having multiple unique DAGs from the same MEC. This shows that while DiBS has a high diversity in terms of MECs (mainly due to covering low-scoring DAGs), DiBS suffers from a lack of diversity with a single MEC, which would be expected from a faithful approximation of the posterior distribution.
\end{itemize}

\end{document}